\documentclass[sigconf]{acmart}

\usepackage{booktabs} 
\usepackage{subcaption}

\usepackage[english]{babel}
\usepackage[utf8x]{inputenc}
\usepackage[T1]{fontenc}
\usepackage{comment} 

\usepackage{amsmath}
\usepackage{graphicx}
\usepackage{xcolor}
\usepackage[colorinlistoftodos]{todonotes}

\copyrightyear{2020}
\acmYear{2020}
\setcopyright{acmcopyright}
\acmConference[GECCO '20]{Genetic and Evolutionary Computation Conference}{July 8--12, 2020}{Cancún, Mexico}
\acmBooktitle{Genetic and Evolutionary Computation Conference (GECCO '20), July 8--12, 2020, Cancún, Mexico}
\acmPrice{15.00}
\acmDOI{10.1145/3377930.3389822}
\acmISBN{978-1-4503-7128-5/20/07}

\begin{document}

\title[CPPN2GAN: Combining CPPNs and GANs for Large-scale Pattern Generation]{CPPN2GAN: Combining Compositional Pattern Producing Networks and GANs for Large-scale Pattern Generation}

\author{Jacob Schrum}
\orcid{0000-0002-7315-0515}
\affiliation{%
  \institution{Southwestern University}
  \streetaddress{1001 E. University Ave}
  \city{Georgetown} 
  \state{Texas, USA} 
  \postcode{78626}
}
\email{schrum2@southwestern.edu}

\author{Vanessa Volz}
\affiliation{%
  \institution{modl.ai}
  \streetaddress{Nørrebrogade 184, 1}
  \city{Copenhagen, Denmark} 
}
\email{vanessa@modl.ai}

\author{Sebastian Risi}
\affiliation{%
  \institution{modl.ai, IT University of Copenhagen}
  \streetaddress{Nørrebrogade 184, 1}
  \city{Copenhagen, Denmark} 
}
\email{sebastian@modl.ai}

\begin{abstract}
Generative Adversarial Networks (GANs) are proving to be a powerful indirect genotype-to-phenotype mapping for evolutionary search, but they have limitations. In particular, GAN output does not scale to arbitrary dimensions, and there is no obvious way of combining multiple GAN outputs into a cohesive whole, which would be useful in many areas, such as the generation of video game levels. Game levels often consist of several segments, sometimes repeated directly or with variation, organized into an engaging pattern. Such patterns can be produced with Compositional Pattern Producing Networks (CPPNs). Specifically, a CPPN can define latent vector GAN inputs as a function of geometry, which provides a way to organize level segments output by a GAN into a complete level. This new CPPN2GAN approach is validated in both \emph{Super Mario Bros.} and \emph{The Legend of Zelda}. Specifically, divergent search via MAP-Elites demonstrates that CPPN2GAN can better cover the space of possible levels. The layouts of the resulting levels are also more cohesive and aesthetically consistent.
\end{abstract}

%
%



\maketitle

\section{Introduction}
Generative Adversarial Networks (GANs \cite{goodfellow2014generative}), which are a type of generative neural network that are trained in an unsupervised way, have been shown capable of reproducing certain aspects of a given training set. For example, they can generate diverse high-resolution samples of a variety of different image classes \cite{brock2018large}.

However, it is an open question how such a GAN-based approach can scale to generate arbitrarily large artefacts that have a modular structure, such as complete game levels. Game levels can consist of many segments repeated many times, often with variation.
Several recent works have shown that it is possible to learn the structure of video game levels using GANs \cite{volz:gecco2018,park:cog19,torrado2019bootstrapping,gutierrez2020zeldagan}, but these approaches only generate  small level segments. 

In particular, Volz et al.~\cite{volz:gecco2018} trained a GAN on small segments of 
Super Mario Bros.\ levels
and searched the induced latent space with CMA-ES \cite{hansen2001ecj:cmaes} to find segments with different properties. In effect, the GAN learned a compact and robust \emph{genotype-to-phenotype} mapping, in which most generated artefacts resemble valid level segments.  
Training such a GAN on complete Mario levels instead of segments would be challenging due to the drastic
reduction in available samples, and the large variation in scale across different levels. While some GAN-based approaches exist to generate complete levels for e.g.\ the game DOOM  \cite{giacomello:cog19}, 
they only encode the high-level map structure without low-level details. Instead, this paper aims to generate complete levels with low-level details, by generating multiple segments with a GAN, and combining them into a cohesive global pattern.
For example, different segments might increase smoothly in difficulty, have repeating structures, or certain motifs might appear in a symmetric fashion. 

\begin{figure*}[t]
\centering
\includegraphics[width=1.0\textwidth]{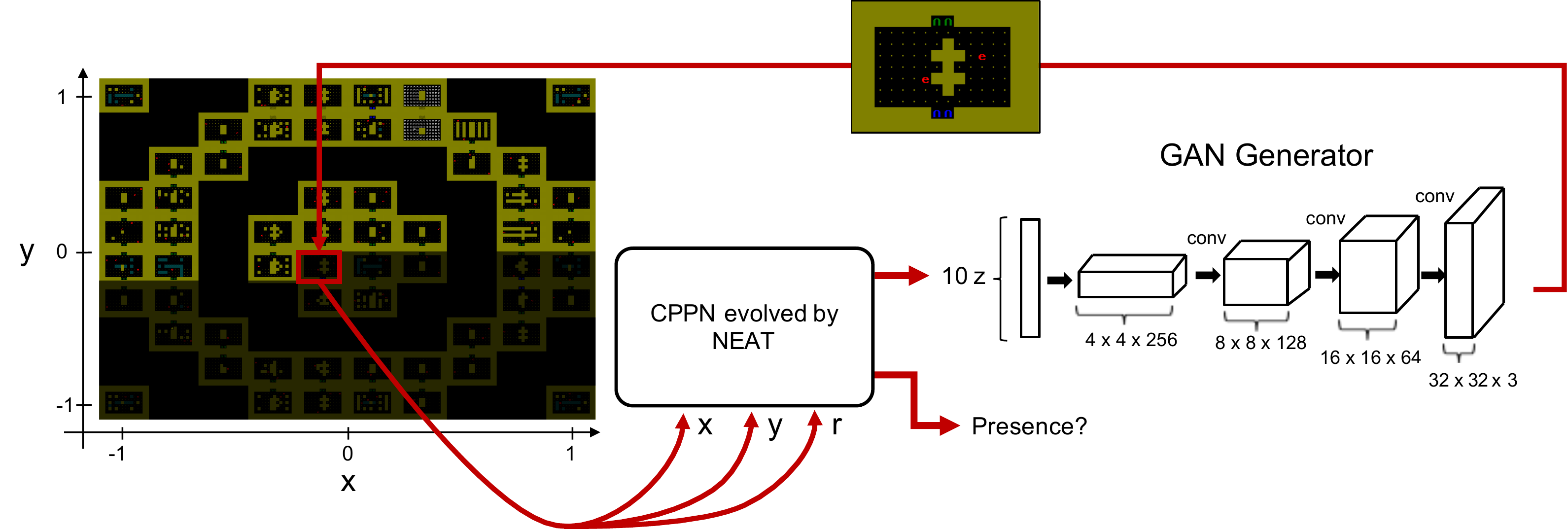}
\caption{CPPN2GAN approach applied to Zelda. \normalfont The CPPN takes as input the Cartesian coordinates of a segment in a level ($x$ and $y$) together with its distance from the center $r$, and for each segment produces a different latent vector $z$ that is then fed into the generator of a GAN pre-trained on existing level content. 
The CPPN also outputs additional information determining whether the room should be placed, how its doors connect to other rooms, and other miscellaneous information.
The approach is able to capture the patterns in the individual level segments, but also create complete maps with global structures such as imperfect radial symmetry in the example here.}
\label{fig:overview}
\end{figure*}

This paper builds on a method that has proven its ability to generate patterns with regularities such as symmetry, repetition, and repetition with variation: Compositional Pattern Producing Networks (CPPNs \cite{stanley:gpem2007}). A CPPN is a special type of neural network that compactly describes patterns with regularities and has succeeded in a variety of domains \cite{secretan:ecj2011,hoover2012generating,clune:ecal11,cellucci20171d,hastings2009automatic,risi2012combining,tweraser:gecco2018}. The main contribution of this paper is a way of combining one GAN and one CPPN: evolve a CPPN that takes Cartesian coordinates of level segments as input, and produces latent vectors for a GAN that was pre-trained on existing level content (Fig.~\ref{fig:overview}). This combination of technologies can generate large-scale and regular video game maps.

Experiments in this paper apply the new technique, CPPN2GAN, to the games Super Mario Bros.\ and The Legend of Zelda\footnote{Code available at \url{https://github.com/schrum2/GameGAN}}. Data from the Video Game Level Corpus \cite{summerville:vglc2016} is used to train GAN models for each game, and CPPNs are evolved using these GANs.
CPPN2GAN is compared to the state-of-the-art for GAN-based level generation, which is the evolution of real-valued genomes consisting of multiple GAN latent vectors, one per segment. This control method is called Direct2GAN. 
Both approaches use the Quality Diversity algorithm MAP-Elites \cite{mouret:arxiv15} to cover the space of possible level designs.
Results below show that Direct2GAN is significantly inferior to the new CPPN2GAN approach in terms of coverage of level design space, and also does not scale to arbitrary level sizes like CPPN2GAN.

Ultimately, the new CPPN2GAN approach could be relevant not only for game levels, but for other domains requiring large-scale pattern generators, such as texture generation, building architecture, neural architecture search, or computer-aided design in general.

\section{Related Work}

This paper combines Generative Adversarial Networks (GANs) and Compositional Pattern Producing Networks (CPPNs) into a new form of Latent Variable Evolution (LVE).

\subsection{CPPNs} 
\label{sec:cppnsAndNEAT}

Compositional Pattern Producing Networks (CPPNs \cite{stanley:gpem2007}) are artificial neural networks (ANNs) with varying activation functions per node. Unlike standard ANNs,
they are repeatedly queried across a geometric space of possible inputs. For example, a CPPN can generate a 2D image by taking Cartesian pixel coordinates $(x,y)$ as input and outputting intensity values for each corresponding pixel.


CPPNs typically include a large array of different activation functions that are biased towards specific patterns and regularities. For example, a Gaussian function allows a CPPN output
pattern to be symmetric, and including a periodic function such as sine can lead to repeating patterns. Other patterns, such as repetition with variation (e.g.\ the fingers of the human hand) can be created by combining functions (e.g.\ sine and Gaussian). CPPNs have been adapted to produce a variety of patterns in domains such as 2D images \cite{secretan:ecj2011}, musical accompaniments \cite{hoover2012generating}, 3D objects \cite{clune:ecal11}, animations \cite{tweraser:gecco2018}, and  physical robots \cite{cellucci20171d}. In games, CPPNs have been used to create particle effects for weapons \cite{hastings2009automatic} and flowers \cite{risi2012combining}. 


CPPNs are traditionally optimized through NeuroEvolution of Augmenting Topologies (NEAT \cite{stanley:ec02}). NEAT starts with a population of simple neural networks: inputs are directly connected to outputs. Throughout evolution, mutations add nodes and connections. NEAT also allows for efficient crossover between structural components with a shared origin.
The benefit of NEAT is that it optimizes both the neural architecture and weights of the network at the same time. More recently, CPPN-inspired neural networks have also been optimized through gradient descent-based approaches  \cite{ha:iclr2017,fernando2016convolution}.  

While CPPNs can create patterns with complex regularities, training CPPNs to recreate particular images is difficult \cite{woolley2011deleterious}. GANs do not share this weakness, which is why we suggest combining them.


\subsection{Generative Adversarial Networks}
\label{sec:gan}

The training process of Generative Adversarial Networks (GANs \cite{goodfellow2014generative}) can be seen as a two-player adversarial game in which a generator $G$ 
and a discriminator $D$ 
are trained at the same time by playing against each other. The discriminator $D$'s job is to classify samples as being generated (by $G$) or real. The discriminator aims to minimize classification error, but the generator tries to maximize it. 
Thus, the generator is trained to deceive the discriminator by generating samples that are good enough to be classified as genuine. Training ideally reaches a steady state where $G$ reliably generates realistic examples and $D$ is no more accurate than a coin flip.


At the end of training, the discriminator $D$ is discarded, and the generator $G$ is used to produce new, novel outputs that capture the fundamental properties present in the training examples. The input to $G$ is some fixed-length vector from a latent space. 
For a properly trained GAN, randomly sampling vectors from this space should produce outputs that could pass as real images or levels. However, to find content with certain properties (such as a specific game difficulty, number of enemies), the latent space needs to be searched, as described next.


\subsection{Latent Variable Evolution}

The first latent variable evolution (LVE) approach was introduced
by Bontrager et al.~\cite{bontrager2017deepmasterprint}.  In their work a GAN is trained on a
set of real fingerprint images and then evolutionary search
is used to find latent vectors that match with subjects in the
dataset.
In another paper, Bontrager et al.~\cite{bontrager2018deep} present an interactive evolutionary system, in which users can evolve the latent vectors for
a GAN trained on different classes of objects (e.g.\ faces or shoes). Because the GAN is trained on a specific target domain, it becomes a compact and robust genotype-to-phenotype mapping (i.e.\ most phenotypes resemble valid domain artifacts) and users
were able to guide evolution towards images that closely resembled
given target images. 
This paper introduces the first indirectly encoded LVE approach. Instead of searching directly for latent vectors, parameters for CPPNs are sought. These CPPNs can generate a variety of different latent vectors, conditioned on the locations of level segments.

\section{Video Game Domains}


The games explored in this paper rely on data from the Video Game Level Corpus (VGLC \cite{summerville:vglc2016}). Specifically, GAN models for Super Mario Bros.\ and The Legend of Zelda are trained, though in each case some specialized processing of the data is required.

\begin{table}[t!]
\centering
\caption{\label{tab:mariotiles}Tile types used in generated Mario levels. \normalfont 
Symbol characters (\textit{sym}) come from the modified VGLC encoding, and the
numeric identity values (\textit{num}) are then mapped to the corresponding values
employed by the Mario AI framework for visualization (\textit{vis}).
The numeric identity values are 
expanded into one-hot vectors when input into the
discriminator network during GAN training.}
\begin{tabular}{lcccl}
\hline
Tile type & sym & num & vis&\\
\hline
Stone & X & 0 & \includegraphics[scale=0.5]{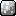}&\\
Breakable & x & 1 & \includegraphics[scale=0.5]{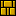}&\\
Empty (passable) & - & 2 & \\
Question Block with coin & q & 3 & \includegraphics[scale=0.5]{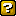}&\\
Question Block with power up & Q & 4 & \includegraphics[scale=0.5]{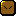}&\\
Coin & o & 5 & \includegraphics[scale=0.5]{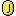}&\\
Pipe & t & 6 & \includegraphics[scale=0.5]{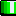}\includegraphics[scale=0.5]{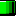}&\\[-6.5px]
& & & \includegraphics[scale=0.5]{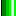}\includegraphics[scale=0.5]{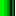}&\\
Piranha Plant Pipe & p & 7 & \includegraphics[scale=0.5]{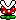}&\\[-6.5px]
 &  &  & \includegraphics[scale=0.5]{encoding_6.png}\includegraphics[scale=0.5]{encoding_7.png}&\\[-6.5px]
& & & \includegraphics[scale=0.5]{encoding_8.png}\includegraphics[scale=0.5]{encoding_9.png}&\\
Bullet Bill & b & 8 & \includegraphics[scale=0.5]{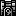}& \hspace{-15px} \includegraphics[scale=0.5]{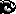}\\[-6.5px]
 &  &  & \includegraphics[scale=0.5]{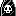}&\\[-6.5px]
 &  &  & \includegraphics[scale=0.5]{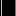}&\\
Goomba & g & 9 & \includegraphics[scale=0.5]{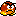} \includegraphics[scale=0.5]{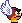}&\\
Green Koopa & k & 10 & \includegraphics[scale=0.5]{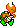} \includegraphics[scale=0.5]{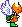} & \\
Red Koopa & r & 11 & \includegraphics[scale=0.5]{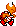} \includegraphics[scale=0.5]{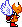} &\\
Spiny & s & 12 & \includegraphics[scale=0.5]{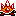} \includegraphics[scale=0.5]{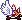} &\\
\hline
\end{tabular}
\end{table}

\subsection{Super Mario Bros.}
\label{sec:mario}

Super Mario Bros.~(1985) is a platform game that involves moving left to right while running and jumping. Levels are visualized with the 
Mario AI framework\footnote{http://marioai.org/}. 


The tile-based level representation comes from VGLC, which uses a particular character symbols to represent each possible tile type. The encoding is extended to more accurately reflect the data in the original game (Table \ref{tab:mariotiles}). For example, VGLC does not distinguish between different enemy types.
Besides adding symbols for more enemy types, the representation of \emph{pipes} is adjusted to avoid the broken \emph{pipes} seen in previous work \cite{volz:gecco2018}.
Instead of using four different tile types for a \emph{pipe}, a single tile is used as an indicator for the presence of a \emph{pipe} and extended automatically downward as required. A detailed explanation of all modifications made to the encoding can be found in work by Volz \cite[Chap. 4.3.3.2]{Volz19}. 


\subsection{The Legend of Zelda}
\label{sec:zelda}

The Legend of Zelda (1986) is an action-adventure dungeon crawler. The main character, Link, explores several dungeons full of enemies, traps, and puzzles. This game also has a tile-based VGLC description. In this paper, the game is visualized  with an ASCII-based Rogue-like game engine used in previous work \cite{gutierrez2020zeldagan}. The mapping between original game tiles and Rogue-like tiles is in Table~\ref{tab:zeldatiles}.

\begin{table}[t!]
\centering
\caption{\label{tab:zeldatiles}Tile types used in generated Zelda rooms. \normalfont
The symbol characters (\emph{sym}) come from the VGLC encoding, and the corresponding
tile images from the original Legend of Zelda are also shown (\emph{game}). 
The diversity of VGLC tile types is mapped down to a smaller set of tiles. The numeric values for GAN training are shown (\emph{num}), and the final tiles depict how they appear in the Rogue-like game engine used for visualization (\emph{rogue}).}
\begin{tabular}{ccccc}
\hline
Tile type & sym & game & num & rogue\\
\hline
Floor & F & \includegraphics[scale=0.75]{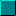} & 0 & \includegraphics[scale=0.75]{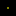} \\
Wall & W & \includegraphics[scale=0.375]{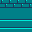} & 1 & \includegraphics[scale=0.75]{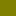} \\
Block & B & \includegraphics[scale=0.75]{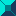} & 1 & 
\includegraphics[scale=0.75]{rouge-wall.png}\\
Door & D & \includegraphics[scale=0.375]{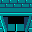} & 1 & \includegraphics[scale=0.75]{rouge-wall.png}\\
Stair & S & \includegraphics[scale=0.75]{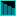} & 1 & \includegraphics[scale=0.75]{rouge-wall.png}\\
Monster statue & M & \includegraphics[scale=0.75]{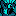} & 1 & \includegraphics[scale=0.75]{rouge-wall.png}\\
Water & P & \includegraphics[scale=0.75]{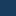} & 2 & 
\includegraphics[scale=0.75]{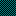}\\
Walk-able Water & O & \includegraphics[scale=0.75]{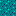} & 2 & \includegraphics[scale=0.75]{rouge-water.png}\\
Water Block & I & \includegraphics[scale=0.75]{block.png} & 2 & \includegraphics[scale=0.75]{rouge-water.png}\\

\hline
\end{tabular}
\end{table}


Previous work \cite{gutierrez2020zeldagan} reduced the large set of tiles inherent to Zelda to a smaller set based on functional requirements. Some Zelda tiles differ in purely aesthetic ways, and others rely on complicated mechanics not implemented in the Rogue-like. 
The Rogue-like accommodates three tile types: a floor tile which directly corresponds to the Zelda floor tile, an impassable tile that corresponds to all impassable tiles in Zelda, and a water tile that corresponds to an obstacle that enemies can pass, but Link cannot.


Enemies are not represented in the VGLC data because its authors did not include them\footnote{VGLC erroneously refers to statues that occupy some rooms as enemies, but they are simply impassable objects. Other enemies are absent from VGLC}. 
VGLC does include information about doors linking rooms, but that information is excluded from the encoding because door placement is handled not by the GAN, but by CPPNs, as described in Section \ref{sec:cppn2gan}.


\section{Approach}

The novel approach introduced in this paper employs a two-stage indirect encoding to evolve video game levels. Individual level segments are created by sending latent vectors to a Generative Adversarial Network (GAN) trained on data for a target game. To create whole levels, locations of individual segments are used as input to a Compositional Pattern Producing Network (CPPN), which outputs latent vectors for each segment. An overview of the complete CPPN2GAN approach below is shown in Fig.~\ref{fig:overview}. For comparison, the Direct2GAN control approach that generates levels from a genome consisting of many separate latent vectors is also described below. 


\subsection{GAN Training Details}
\label{sec:processing}

The Mario model used was taken from a publicly
available repository associated with previous research \cite{volz:gecco19}, 
but details of its training are repeated here.
The Zelda model is new, but is 
based on another model from a recent paper \cite{gutierrez2020zeldagan}.
Both models are Wasserstein GANs \cite{arjovsky2017wasserstein} differing
only in the size of their latent vector inputs (10 for Zelda, 30 for Mario), and the depth of the final output layer (3 for Zelda, 13 for Mario).
Their architecture otherwise matches that shown in Fig.\ \ref{fig:overview}.
Output depth corresponds to the number of possible tiles for the game. The other output dimensions are $32 \times 32$, which is larger than the 2D region needed to render a level segment. During training and generation, the upper left corner of the output region is treated as a generated level, and the rest is ignored.



To encode the levels for training,
each tile type is represented by
a distinct integer, which is converted to a one-hot
encoded vector before being input into the
discriminator. 
The generator also outputs
levels in 
the one-hot encoded format, which is then converted
back to a collection of integers.
Mario or Zelda levels in this integer-based format can be sent to
the Mario AI framework or Rogue-like engine for rendering.
The mapping from VGLC tile types and symbols, 
to GAN training number codes, and finally to Mario AI/Rogue-like 
tile visualizations
is detailed in Tables~\ref{tab:mariotiles} and~\ref{tab:zeldatiles}. 

GAN input files for Mario were created by processing all 12 \emph{overworld} level files
from VGLC for Super Mario Bros.
Each level is 14 tiles high. The GAN expected to always see a rectangular
input of the same size, so each input was generated
by sliding a 28 (wide) $\times$ 14 (high)
window over the level from left to right, one tile at a time. The width of 28 tiles is the width of the screen in Mario. 



The GAN input for Zelda was created from the 18 dungeon files in VGLC for The Legend of Zelda, but the actual training samples are the individual rooms in the dungeons, which are 16 (wide) $\times$ 11 (high) tiles in size. Many rooms are repeated within and across dungeons, so only unique rooms were included in the training set (only 38 samples). The training samples are simpler than the raw VGLC rooms because the various tile types are reduced to a set of just three as shown in Table~\ref{tab:zeldatiles}. Doors were transformed into walls because door placement is not handled by the GAN, but rather by the CPPN2GAN approach, described next.

\subsection{Level Generation: CPPN2GAN}
\label{sec:cppn2gan}


In order to generate a level using CPPNs and GANs, the CPPN is given responsibility for generating latent vector inputs for the GAN as a function of segment position within the larger level.

For Mario, the only input is the x-coordinate of the segment, scaled to $[-1,1]$. So, for a level of three segments, the CPPN inputs would be $-1$, $0$, and $1$. The output of the CPPN is an entire latent vector (30 variables for Mario). Each latent vector is fed to the GAN to generate the segment at that position in the level.

Zelda's 2D arrangement of rooms is more complicated. 
For the overall dungeon shape to be interesting, some rooms in the 2D grid need to be missing. 
Also, dungeons are typically more interesting if they are maze-like, so simply connecting all adjacent rooms would be boring. How maze-like a dungeon is also depends on its start and end points.
These additional issues are global design issues, and therefore are handled by the CPPN, which defines global patterns, rather than the GAN, which generates individual rooms.

Thus, CPPNs for Zelda generate latent vector inputs and additional values that determine the layout and connectivity of the rooms.
Zelda CPPNs take inputs of x and y coordinates scaled to $[-1,1]$. A radial distance input is also included to encourage radial patterns, which is common in CPPNs \cite{secretan:ecj2011}. For each set of CPPN inputs, the output is a latent vector (size 10 for Zelda) along with six additional numbers: room presence, right door presence, down door presence, right door type, down door type, and start/end preference.

Room presence determines the presence/absence of a room based on whether the number is positive. Similarly, if a room is present and has a neighboring room in the given direction, 
then positive right/down door presence values place a door in the wall heading right/down. 
Whenever a door is placed, a door is also placed in the opposite direction within the connecting room, which is why top/left door outputs are not needed. For variety, the right/down door type determines the types of doors, based on different number ranges for each door type: $[-1,0]$ for plain, $(0,0.33]$ for soft-locked, $(0.33,0.66]$ for bomb-able passage, and $(0.66,1.0]$ for locked. Soft-locked doors only open when all enemies in the room are killed, bomb-able passages are secret walls that can be bombed to create a door, and locked doors need a key. Enough keys to pair with all locked doors are placed at random locations in rooms of the dungeon. However, to assure that the genotype to phenotype mapping is deterministic, the pseudo-random generator responsible for placing keys is initialized using the bit representation of the corresponding right or down door type output as a seed.

The final output for start/end preference determines which rooms are the start/end rooms of the dungeon. Across all rooms in the dungeon, the one whose start/end preference is smallest is the player's starting room, and the one with the largest output is the final goal room, designated by the presence of a Triforce item (triangle).

This new approach to generating complete levels is compared with the control approach described next.

\subsection{Level Generation: Direct2GAN}
\label{sec:direct2gan}

To have a meaningful comparison with the CPPN2GAN approach, an approach that directly evolves genomes consisting of multiple latent vectors is needed. The Direct2GAN method  evolves levels consisting of $S$ segments for a GAN expecting latent inputs of size $Z$ by evolving real-valued genome vectors of length $S \times Z$. Each genome is chopped into individual GAN inputs at level generation.

This approach requires a convention as to how different segments are combined into one level. For Mario's linear levels, 
adjacent GAN inputs from the combined vector correspond to adjacent segments in the generated level. The combined vector is simply processed from left to right to produce segments from left to right.

To generate 2D Zelda dungeons, individual segments of the linear genome are 
mapped to a 2D grid in row-major order: processing genome from left to right generates top row from left to right, then moves to next row down and so on. For fair comparison with CPPN2GAN, each portion of a genome corresponding to a single room contains not only the latent vector inputs, but the six additional numbers for controlling global structure and connectivity: room presence, right door presence, down door presence, right door type, down door type, and start/end preference. Therefore, a $M \times N$ room grid requires genomes of length $M \times N \times (Z + 6)$. 

Such massive genomes induce large search spaces that are difficult to search, as demonstrated by the experiments described next.

\section{Experiments}


Demonstrating the expressive range of new game level encodings is important.
Therefore, the Quality Diversity algorithm MAP-Elites \cite{mouret:arxiv15}, which divides the search space into phenotypically distinct bins, is used for evolving both CPPNs and real-valued vectors.

\subsection{MAP Elites}

Instead of only optimizing towards an objective, as in standard evolutionary algorithms, MAP Elites (Multi-dimensional Archive of Phenotypic Elites \cite{mouret:arxiv15})  collects a diversity of quality artefacts that differ along a number $N$ of predefined dimensions. MAP-Elites discretizes the space of produced artefacts into different bins and, given some objective, keeps track of the highest performing individual for each bin in the $N$-dimensional behavior space. 

Our implementation starts by generating an initial population of 100 random individuals that are placed in bins based on their attributes. Each bin only holds one individual, so individuals with higher fitness replace less fit individuals. Once the initial population is generated, solutions are randomly sampled uniformly from the bins and undergo crossover and/or mutation to generate new individuals. These newly created individuals also replace less fit individuals as appropriate, or end up occupying new bins, filling out the range of possible designs.
Performance is measured both in terms of achieved fitness and the number of bins that are filled as more individuals are generated. Our experiments generate 50,000 individuals per run after the initial population is generated.


To support a range of meaningful variation within each domain, each game uses its own distinct binning scheme and fitness measure.

\subsection{Dimensions of Variation Within Levels}


In Mario, the bin dimensions are based on measurements of three quantities: decoration frequency, space coverage, and leniency. These measures were inspired by a study on evaluation measures for platformer games \cite{Summerville2017}. 
Each measure expresses different characteristics of a level:
\begin{itemize}
    \item decoration frequency: Percentage of \emph{non-standard} tiles\footnote{breakable tiles, question blocks, pipes, all enemies} 
    \item space coverage: Percentage of tiles 
    that Mario can stand on\footnote{solid and breakable tiles, question blocks, pipes and bullet bills}
    \item leniency: Average of leniency values\footnote{1: question blocks; -0.5: pipes, bullet bills, gaps in ground; -1: moving enemies; 0: remaining} across all tiles. 
    Enemies and gaps have negative values, power-ups positive values.
\end{itemize}
Each score highlights different aspects of a level. All measures focus on visual characteristics, but also relate to how a player can navigate through a level. These scores were calculated for individual segments (10 per level) and then summed across the segments. 


Preliminary experiments were conducted to uncover reasonable ranges for binning. 
Multiplying decoration frequency and space coverage by 3.0 scales them roughly to the range $[0,1]$, which is evenly split into 10 bins. Only leniency has both negative and positive values. Multiplying the raw score by 5.0 scales it roughly to the range $[-0.5,0.5]$, which is evenly divided into 5 negative bins and 5 positive bins. Negative bins correspond to greater challenges, and positive bins correspond to easier levels. In each dimension, rare out-of-range values were assigned to the nearest bin.



The fitness is the length of the shortest path to beat the level. The objective is to maximize the path length, which favors levels that require jumps - the main mechanic of the game. If no path can be found the level is deemed unsolvable and receives a fitness of 0.

To determine the path, A* search is performed on the tile-based representation of the level. 
To limit computational costs, the game physics are simplified as follows. 
First, Mario always occupies a single tile (small state), and can either move left, move right, or jump. Once a jump starts, Mario moves up one tile per action for the next four actions (unless there is an obstruction), and then falls down until landing on an impassible tile. Mario can still make left/right movements while airborne, but can only initiate a jump if on solid ground. Enemies are just impassible obstacles, and Mario dies if he falls in a pit. The heuristic used favors states further to the right, because the right edge of the screen is the goal.

Variation across Zelda levels is based on water tile percentage, wall tile percentage, and the number of reachable rooms. A room is reachable if it is the start room, or if a door connects it to any reachable room. This definition makes it computationally cheap to determine if a room is \emph{reachable}, but ignores that single rooms can be impassable\footnote{For example because a needed key is inaccessible}.
Water and wall tile percentages are calculated only with respect to reachable rooms, and only for the $12 \times 7$ floor regions of rooms (surrounding walls ignored). Bins for these dimensions are divided into 10\% ranges, creating 10 bins per dimension. However, some of these bins are impossible to fill, because the sum of water and wall percentages must be less than 100\% to fit in the room. 
Floor tiles occupy additional space. For the number of reachable rooms, there is a bin for each possible number out of 100. The maximum number is 100 because dungeons are generated in a $10 \times 10$ grid.

The fitness for Zelda dungeons is the percentage of reachable rooms traversed by the shortest solution path from start to goal. The objective is to maximize the number of rooms visited, as exploring is one of the main mechanics of the game. Again, if no path can be found, the dungeon is deemed unsolvable and receives 0 fitness. 

A* is again used to determine the path, now using Manhattan distance to the goal as a heuristic. We take 
locked doors and keys into account to only generate paths with 
rooms that are \emph{actually} reachable. Since the inclusion of keys makes the state space very large, there is a computation budget of 100,000 states. 


\begin{figure*}[t]
\centering
\begin{subfigure}{0.49\textwidth} 
    \includegraphics[width=1.0\textwidth]{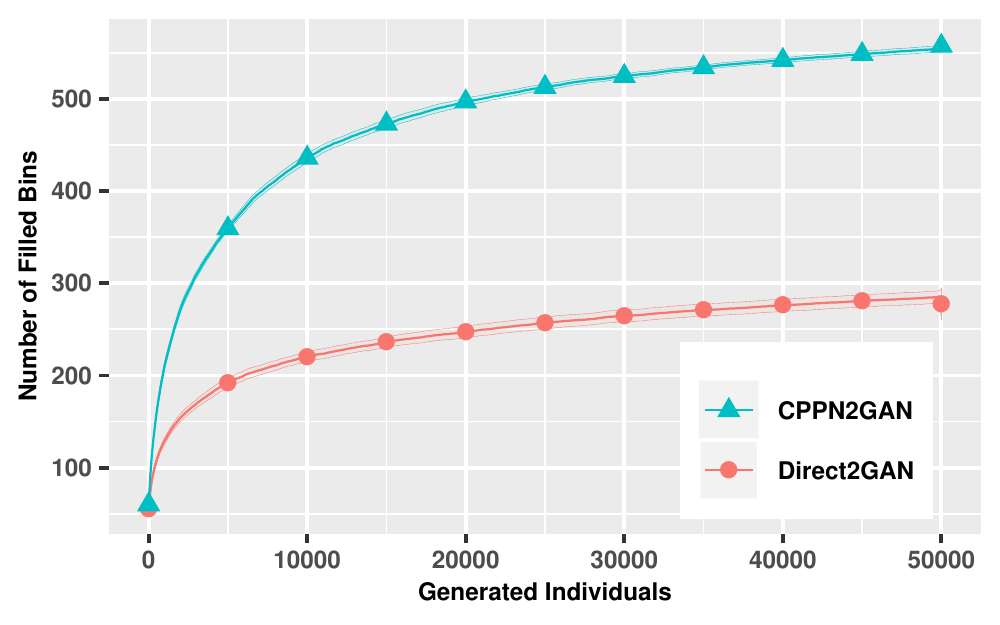}
    \caption{Mario}
    \label{fig:marioBins}
\end{subfigure}
\begin{subfigure}{0.49\textwidth}
    \includegraphics[width=1.0\textwidth]{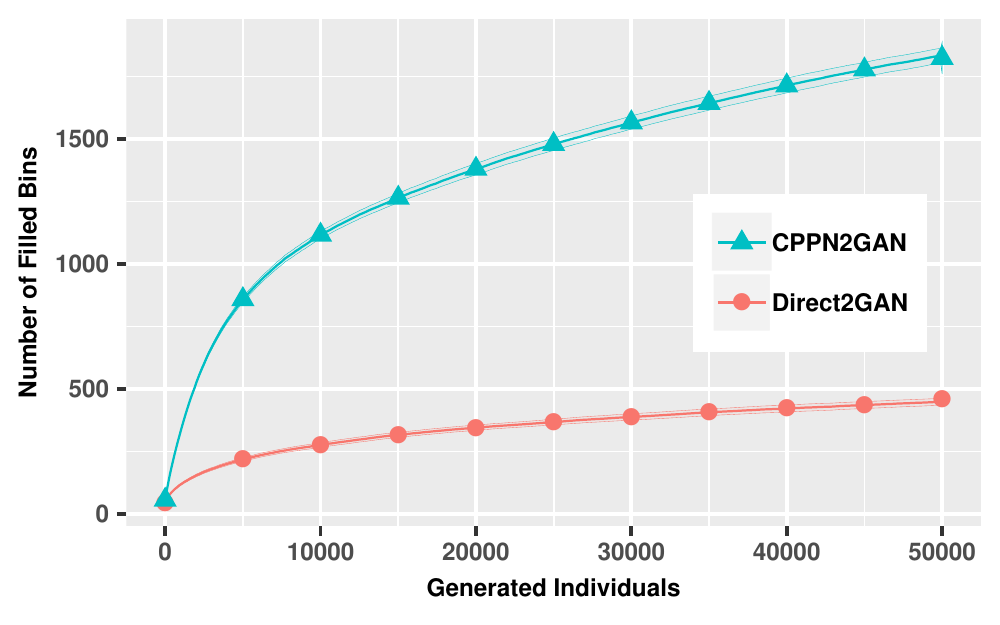}
    \caption{Zelda}
    \label{fig:zeldaBins}
\end{subfigure}
\caption{Average Number of Filled Bins. \normalfont Depicts the average number of bins filled across 30 runs as more levels are generated for both (\subref{fig:marioBins}) Mario, and (\subref{fig:zeldaBins}) Zelda. The 95\% confidence intervals are also shown, and are very narrow, indicating high consistency in performance. CPPN2GAN fills significantly more bins than Direct2GAN in both domains for nearly the entirety of evolution.}
\label{fig:filledBins}
\end{figure*}

\subsection{CPPN Evolution Details}

Levels for both games are evolved using a variant of NEAT \cite{stanley:ec02},
as described in Section \ref{sec:cppnsAndNEAT}. The specific implementation is
MM-NEAT\footnote{\url{https://github.com/schrum2/MM-NEAT}}.
Because CPPNs are being evolved, every neuron in each network can have a different activation function from the following list: 
sawtooth wave, 
linear piecewise, 
id,
square wave, 
cosine
sine, 
sigmoid,
Gaussian, 
triangle wave, and 
absolute value. 

Whenever a new network is generated, is has a 50\% chance of being the offspring of two parents rather than a clone. The resulting network then has a 20\% chance of having a new node spliced in, 40\% chance of creating a new link, and a 30\% chance of randomly replacing one neuron's activation function. There is a per-link perturbation rate of 5\%.

\subsection{Real-Valued Evolution}

For Direct2GAN, real-valued vectors are initialized with random values in the range $[-1,1]$. When offspring are produced, there is a 50\% chance of single-point crossover. Otherwise, the offspring is a clone of one parent. Either way, each real number in the vector then has an independent 30\% chance of polynomial mutation \cite{deb1:cs95:polynomial}.

\section{Results}

\begin{figure*}[t]
\centering
\begin{subfigure}{0.49\textwidth} 
    \includegraphics[width=1.0\textwidth]{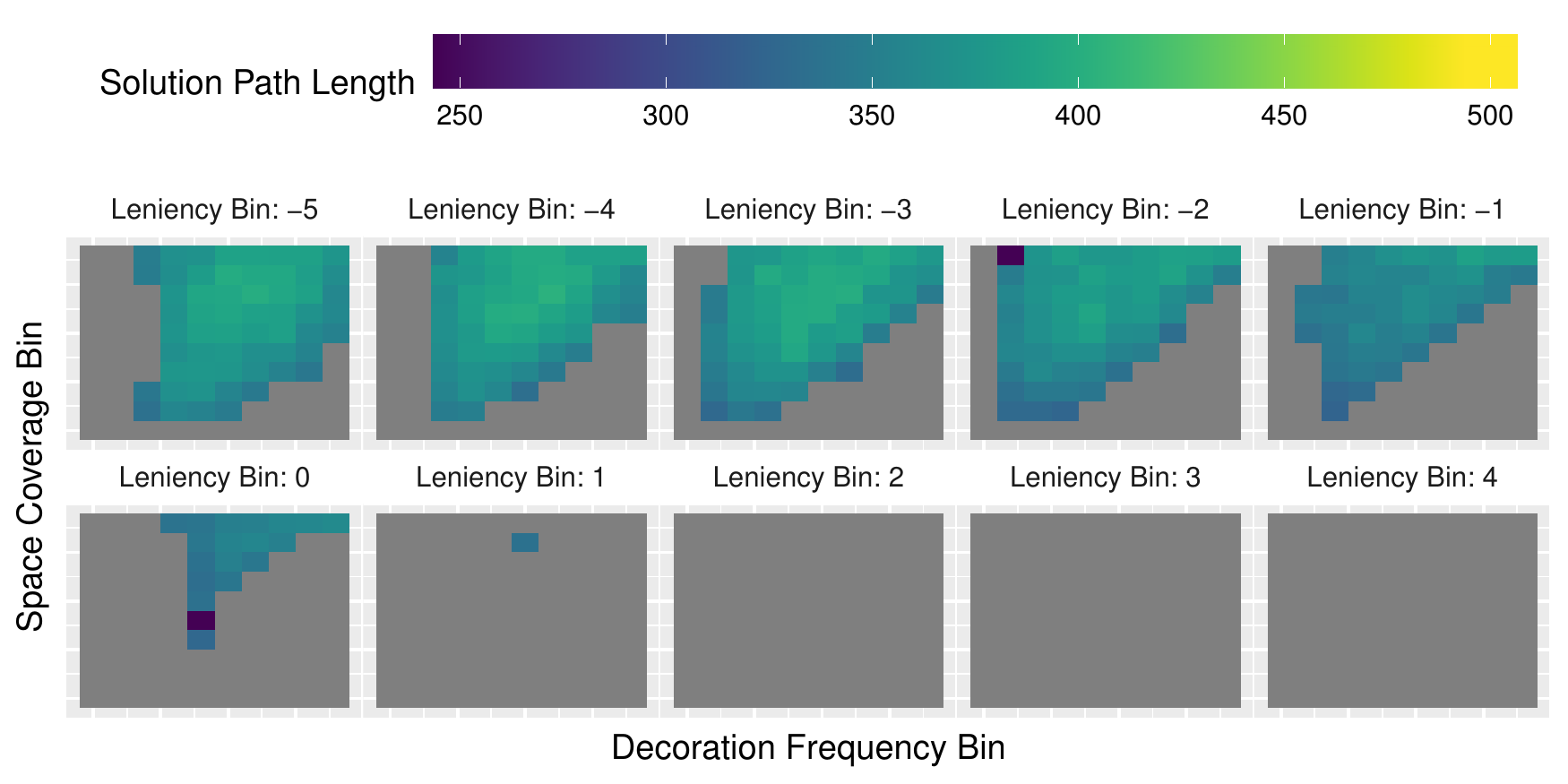}
    \caption{Direct2GAN}
    \label{fig:marioheatDirectToGAN}
\end{subfigure}
\begin{subfigure}{0.49\textwidth}
    \includegraphics[width=1.0\textwidth]{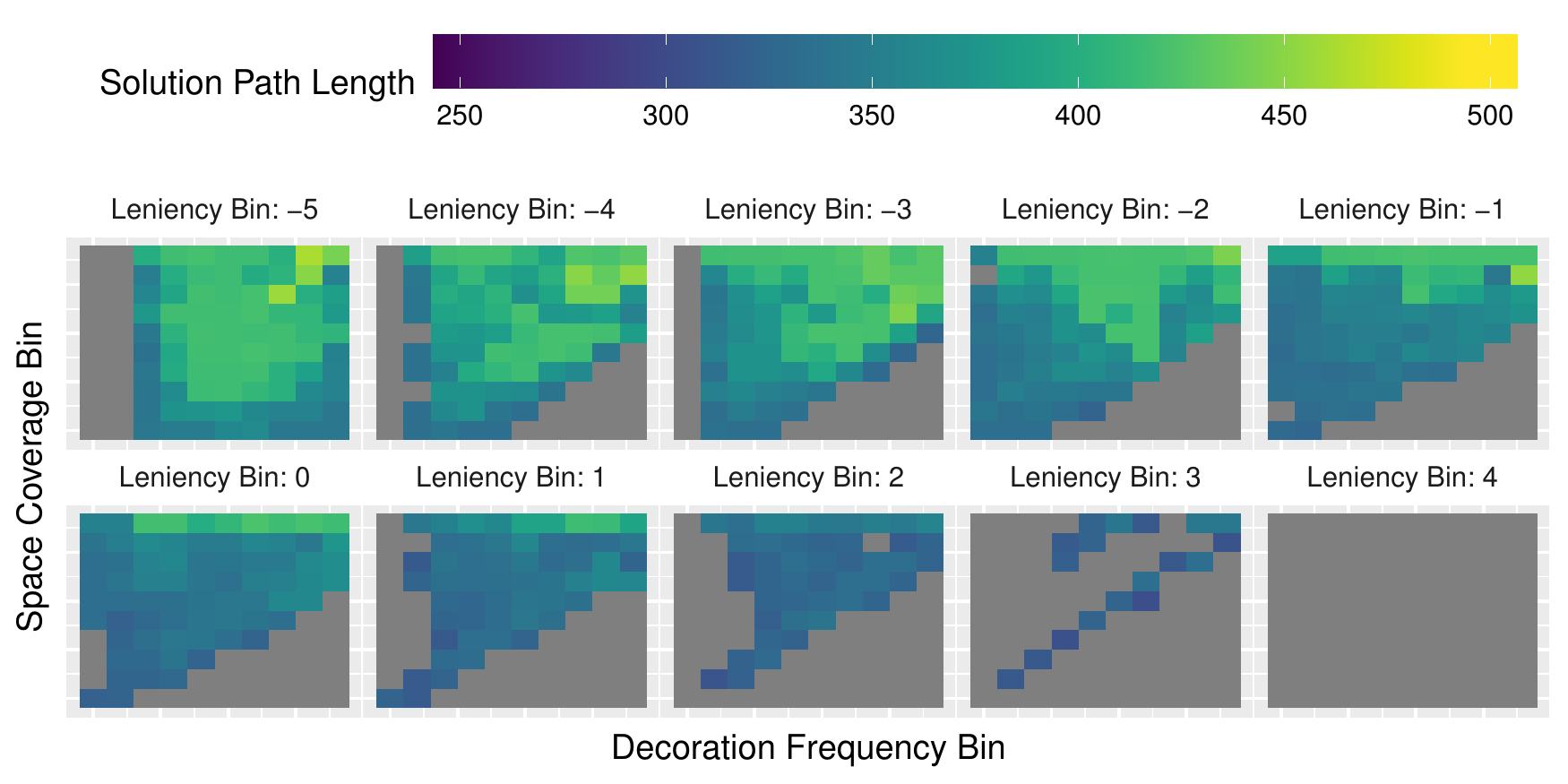}
    \caption{CPPN2GAN}
    \label{fig:marioheatCPPNtoGAN}
\end{subfigure}
\caption{Heatmaps of Typical Final Archives for Mario Level Evolution. \normalfont Final MAP-Elites archives are depicted for one run each of (\subref{fig:marioheatDirectToGAN}) Direct2GAN, and (\subref{fig:marioheatCPPNtoGAN}) CPPN2GAN. CPPN2GAN fills bins with positive leniency that Direct2GAN cannot fill. Both approaches fail to create levels with high decoration and low space coverage, since such features contradict each other. Even in negative leniency regions, CPPN2GAN fills more bins and with higher fitness.}
\label{fig:marioHeatmaps}
\end{figure*}

\begin{figure*}[t]
\centering
\begin{subfigure}{\textwidth} 
    \includegraphics[width=1.0\textwidth]{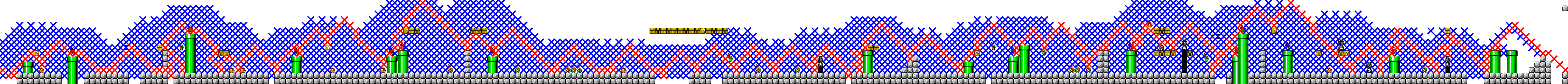}
    \caption{Direct2GAN}
    \label{fig:marioLevelDirectToGAN}
\end{subfigure} \\
\begin{subfigure}{\textwidth}
    \includegraphics[width=1.0\textwidth]{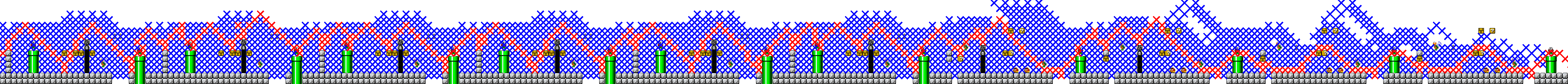}
    \caption{CPPN2GAN}
    \label{fig:marioLevelCPPNtoGAN}
\end{subfigure}
\caption{Example Evolved Mario Levels. \normalfont Each level is shown with \textcolor{blue}{blue X} marks at positions checked by A* search, and a trail of \textcolor{red}{red X} marks along the solution path. Both (\subref{fig:marioLevelDirectToGAN}) Direct2GAN and (\subref{fig:marioLevelCPPNtoGAN}) CPPN2GAN levels are shown. These levels are in the same bin, but from different runs. The CPPN2GAN level begins with one repeating pattern of segments, but then switches to another in the latter half.}
\label{fig:marioLevels}
\end{figure*}

\begin{figure*}[t]
\centering
\begin{subfigure}{0.47\textwidth} 
    \includegraphics[width=1.0\textwidth]{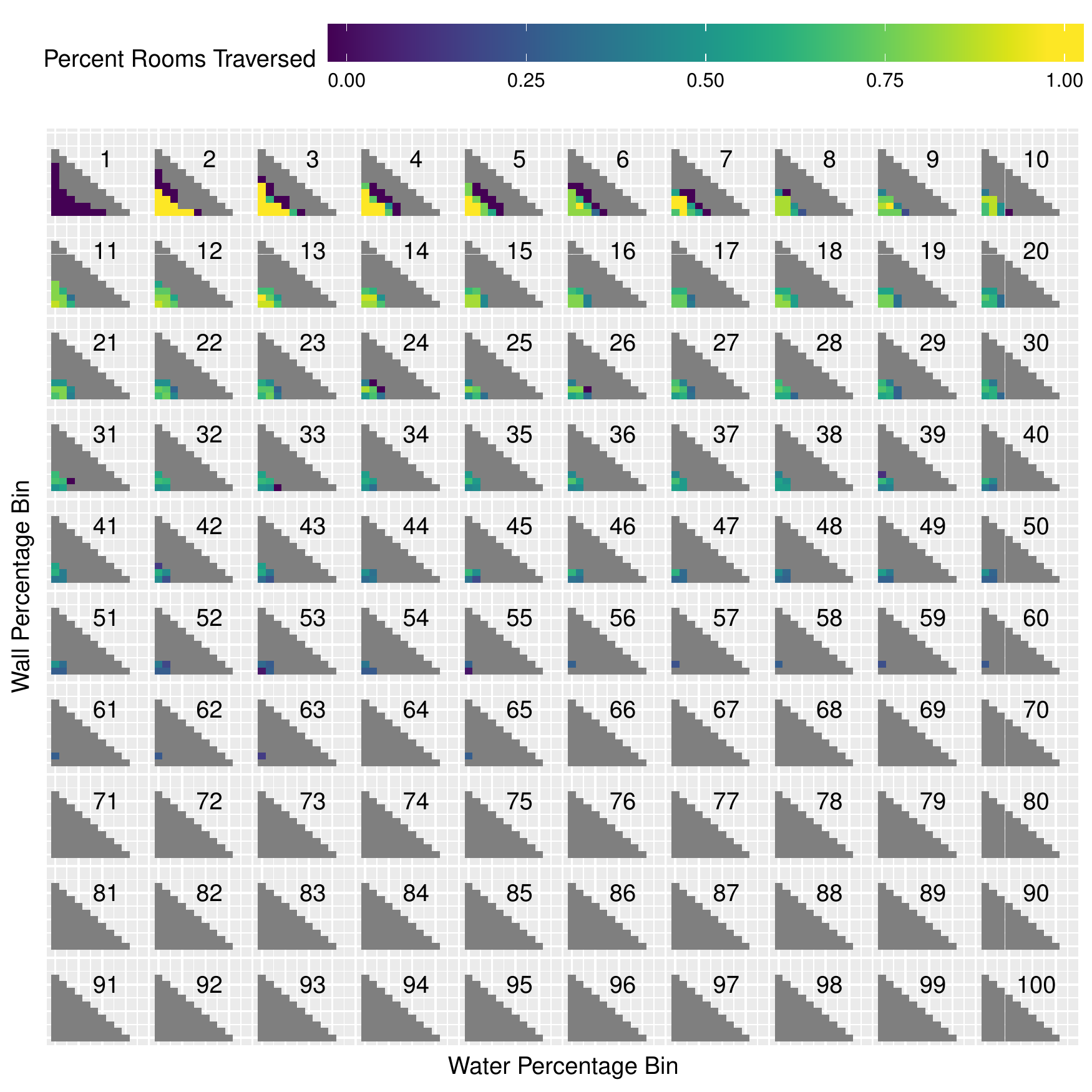}
    \caption{Direct2GAN}
    \label{fig:heatDirectToGAN}
\end{subfigure}
\begin{subfigure}{0.47\textwidth}
    \includegraphics[width=1.0\textwidth]{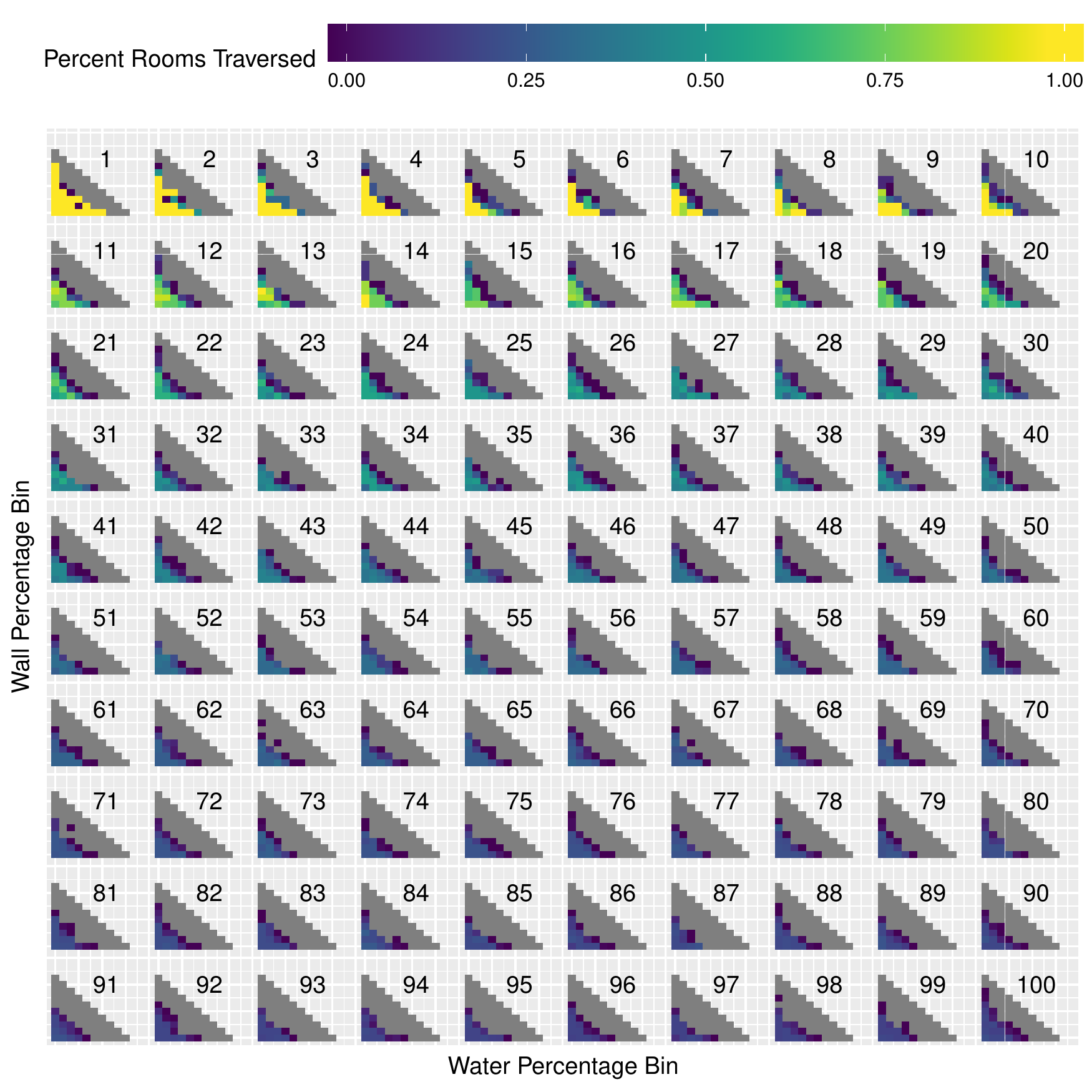}
    \caption{CPPN2GAN}
    \label{fig:heatCPPNtoGAN}
\end{subfigure}
\caption{Heatmaps of Typical Final Archives for Zelda Dungeon Evolution. \normalfont Final MAP-Elites archives are depicted for one run each of (\subref{fig:heatDirectToGAN}) Direct2GAN, and (\subref{fig:heatCPPNtoGAN}) CPPN2GAN.
Each square corresponds to dungeons having a different number of reachable rooms, shown in the upper-right corners. Within each square are 10 columns for the percentage of room tiles that are water, partitioned into groups of 10\%. The rows of each square indicate the percentage of wall tiles in a similar way. The color intensity is the fitness of the individual in the bin: the percent of reachable rooms traversed from start to goal. Note that the upper-right corners of each square are impossible to fill, because the sum of water and wall percentage must be less than 100\%. CPPN2GAN has multiple representatives with every possible number of reachable rooms, but Direct2GAN struggles to create dungeons with larger numbers of reachable rooms.}
\label{fig:heatmaps}
\end{figure*}

\begin{figure*}[t]
\centering
\begin{subfigure}{0.33\textwidth} 
    \includegraphics[width=1.0\textwidth]{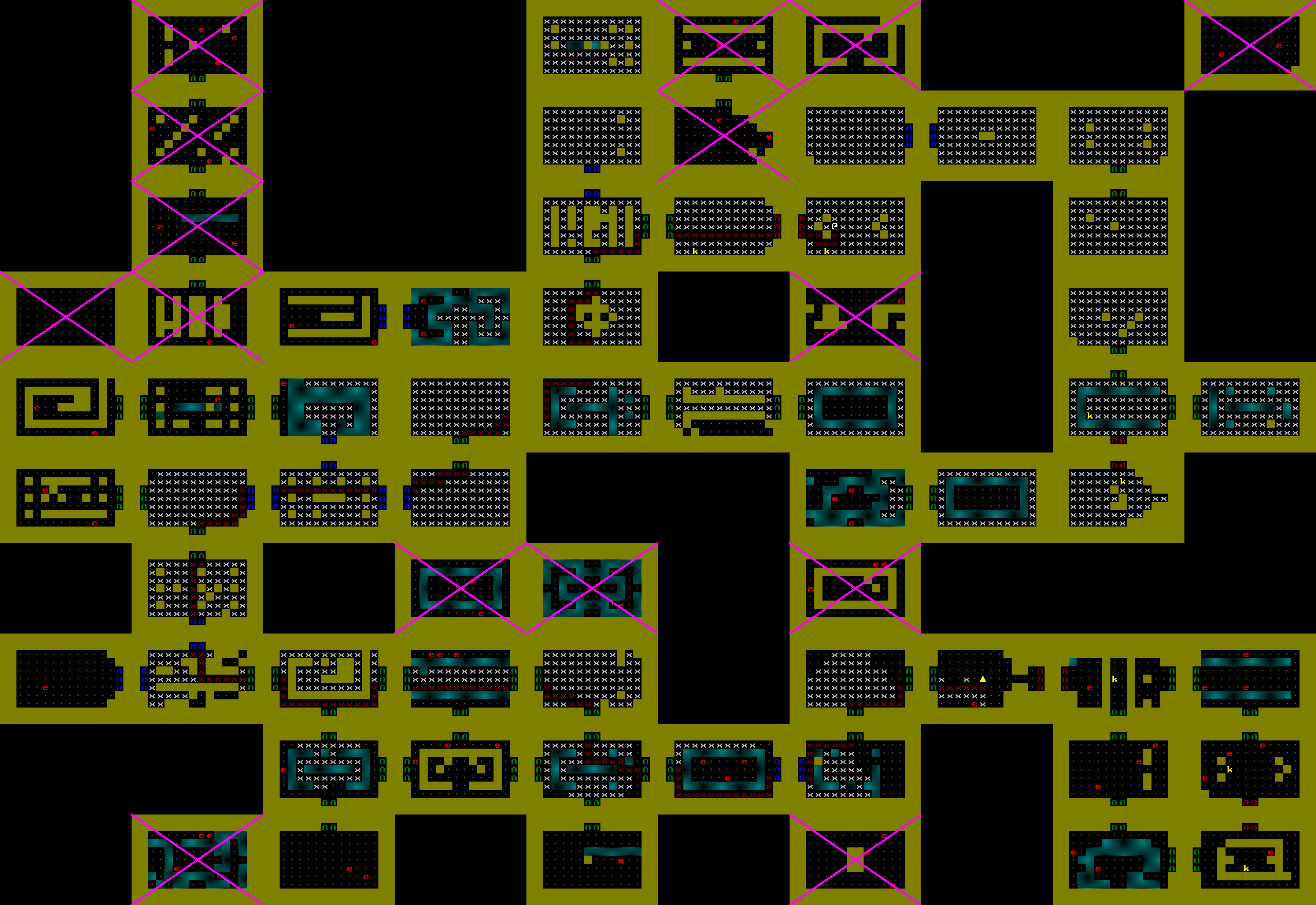}
    \caption{Direct2GAN Dungeon: 50 Reachable Rooms}
    \label{fig:dungeonDirectToGAN}
\end{subfigure}
\begin{subfigure}{0.33\textwidth}
    \includegraphics[width=1.0\textwidth]{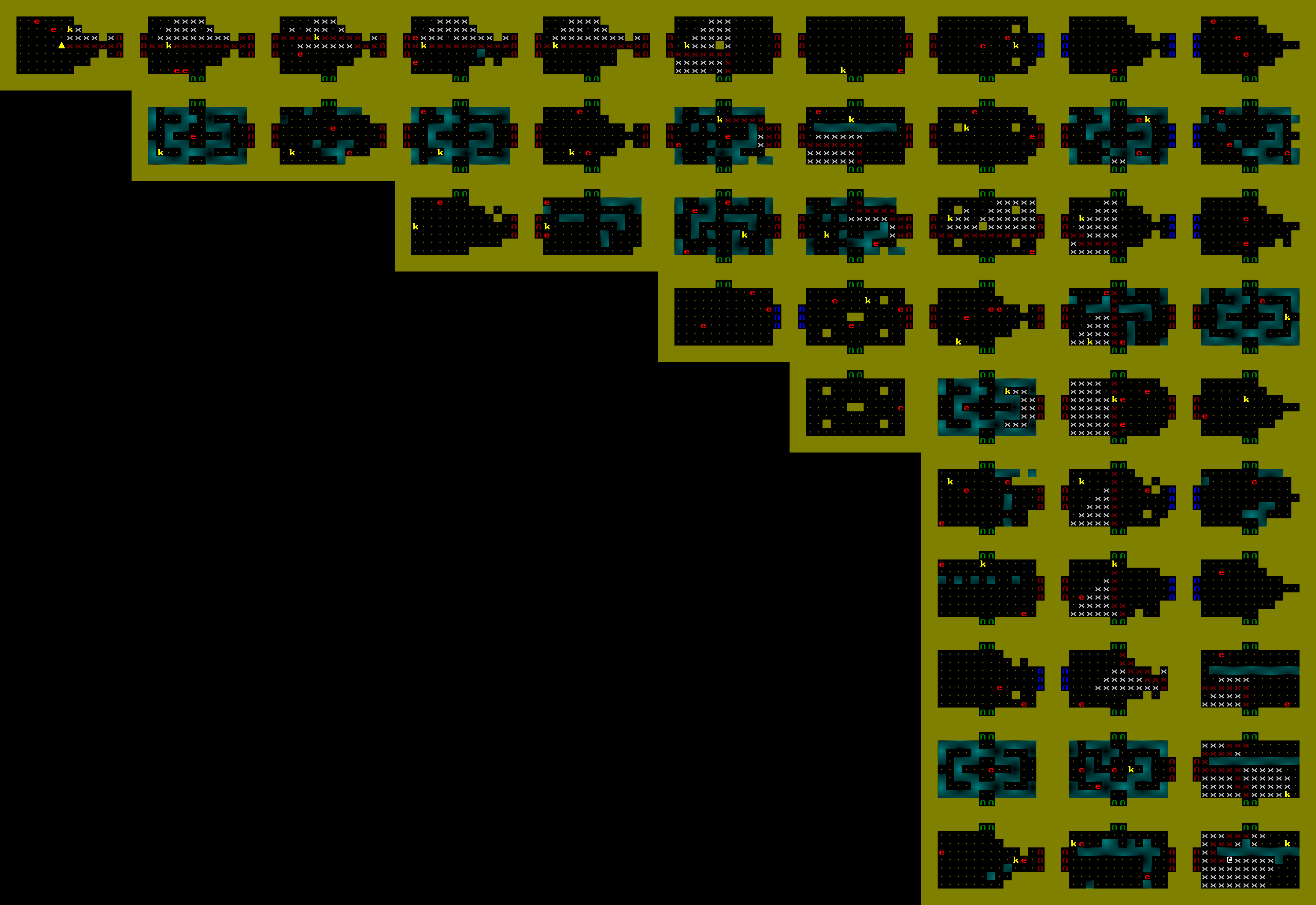}
    \caption{CPPN2GAN: 50 Reachable Rooms}
    \label{fig:dungeonCPPNtoGAN}
\end{subfigure}
\begin{subfigure}{0.33\textwidth}
    \includegraphics[width=1.0\textwidth]{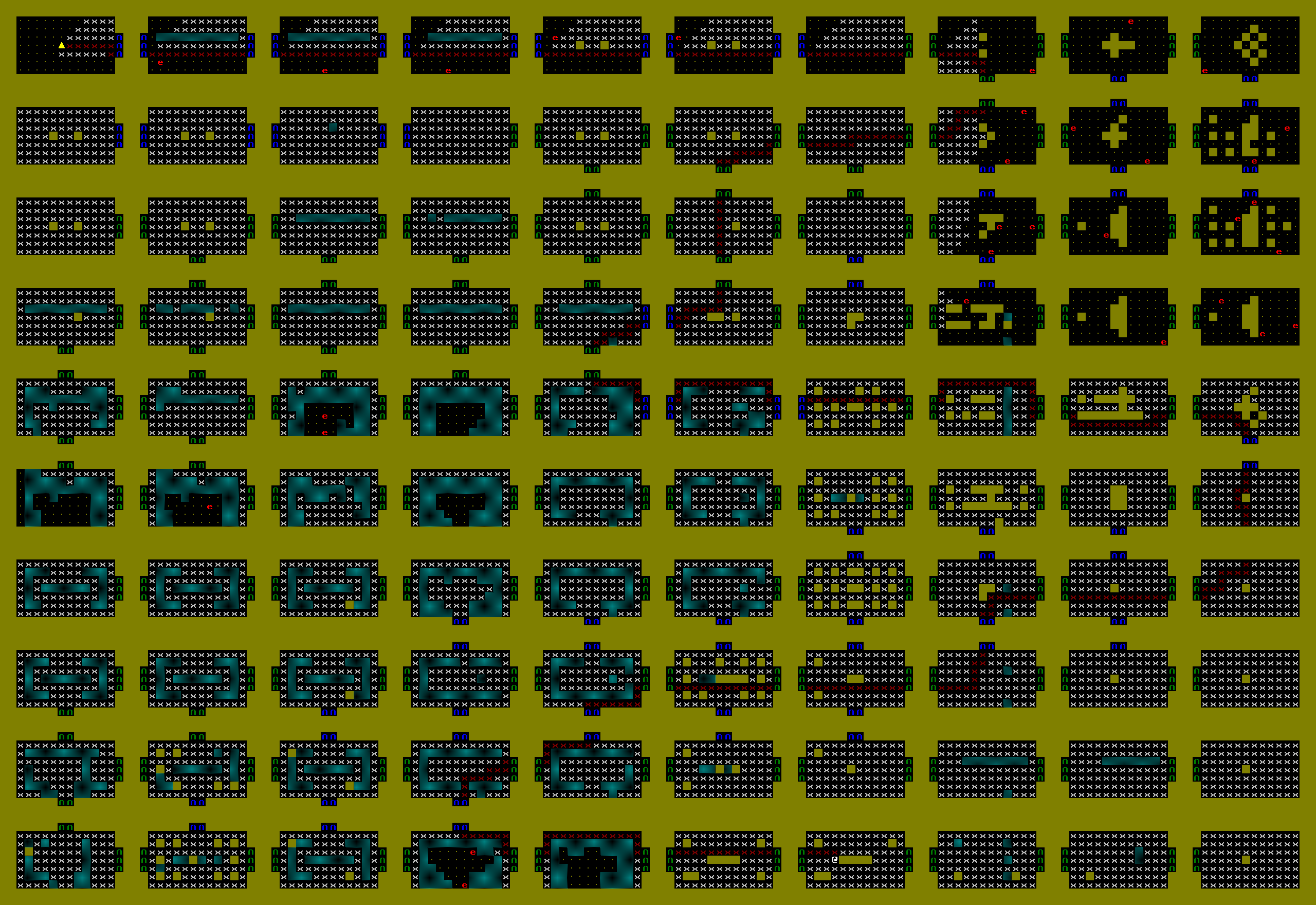}
    \caption{CPPN2GAN: 100 Reachable Rooms}
    \label{fig:dungeonCPPNtoGAN100}
\end{subfigure}
\caption{Example Evolved Dungeons. \normalfont The 
(\subref{fig:dungeonDirectToGAN}) Direct2GAN and (\subref{fig:dungeonCPPNtoGAN}) CPPN2GAN
dungeons with 50 reachable rooms each have the same fitness (38\% of reachable rooms traversed in solution path). There is also a (\subref{fig:dungeonCPPNtoGAN100}) CPPN2GAN dungeon with 100 reachable rooms and a comparable fitness of 34\% reachable rooms. Direct2GAN could not produce dungeons with so many reachable rooms. 
The Direct2GAN dungeon is more sprawling, and has several rooms that are not reachable (\textcolor{magenta}{magenta X}). Both CPPN2GAN dungeons are more cohesive, and themes can be noticed in different regions of the dungeons.
Larger versions of these figures are included in supplementary material.}
\label{fig:dungeons}
\end{figure*}


In terms of the number of MAP Elites bins filled during search, CPPN2GAN quickly surpasses Direct2GAN by a statistically significant amount ($p < 0.05$) in both Mario and Zelda (Fig.\ \ref{fig:filledBins}). The differences are large, and confidence intervals are narrow.

Final heatmaps from individual runs demonstrate which regions of design space each method occupies. Fig.\ \ref{fig:marioHeatmaps} shows final results in Mario for typical runs of Direct2GAN and CPPN2GAN. Direct2GAN seems to have trouble discovering levels with a positive leniency. This is likely because the only way to reach these high leniency values is by adding lots of question blocks and few enemies, which limits the number of suitable latent vectors. Here CPPN2GAN benefits from its encoding, which makes it easier to re-use discovered good solutions. Additionally, even in bins with negative leniency, CPPN2GAN fills in a wider range of decoration frequency and space coverage values, and with higher fitness. Interestingly, both methods have trouble discovering levels with both low space coverage and high decoration frequency. This is because there are several decorative tiles that Mario can stand on (pipes, bullet bills, question blocks, breakable tiles) which thus automatically increase space coverage as well. 
Example evolved levels are in Fig.\ \ref{fig:marioLevels}. 

Example Zelda heatmaps are in Fig.\ \ref{fig:heatmaps}. They reveal that Direct2GAN has trouble creating dungeons with many reachable rooms, whereas CPPN2GAN successfully discovers dungeons with every possible number of reachable rooms. Even for smaller dungeons, CPPN2GAN fills out the range of possible water and wall percentages more fully, and with higher fitness. Example evolved dungeons are in Fig.\ \ref{fig:dungeons}. Direct2GAN has trouble connecting scattered rooms, whereas CPPN2GAN, as expected, lays out rooms in a cohesive pattern with more uniformly structured connectivity.


These experiments demonstrate that CPPN2GAN is vastly superior to Direct2GAN in terms of how much of the design space is covered for both Mario and Zelda. Further, if specific rare level characteristics are desired (e.g.\ Mario levels with high leniency), CPPNs are more likely to discover suitable levels due to indirect encoding. 
However, if diversity within the generated content is desired, Direct2GAN is the better approach. While Mario levels generated with CPPN2GAN do switch between patterns as shown in Fig.\ \ref{fig:marioLevelCPPNtoGAN}, Direct2GAN creates more variation (Fig.\ \ref{fig:marioLevelDirectToGAN}). This of course comes with the cost of losing global level structure.


Similarly, CPPN2GAN is better able to discover Zelda dungeons with specific properties (such as water and wall percentage), 
but CPPN2GAN is better at producing variation in Zelda than in Mario. 
Rooms are still frequently repeated, but not as extensively. Confined geographic regions in each dungeon feature particular tile configurations more prominently, and smoothly transition from one region to another in an organic fashion. This result is especially interesting due its potential correlation with the in-game narrative. For instance, the dungeon in Fig.\ \ref{fig:dungeonCPPNtoGAN} starts in a region of water-filled maze-like rooms, leads the player through rooms with a more open floor plan, then proceeds through a region of different water obstacles before a wide open set of rooms funnels the player to the Triforce. The dungeon in Fig.\ \ref{fig:dungeonCPPNtoGAN100} is interesting because the solution path has several twists and turns, despite all rooms being reachable. This path results from how the CPPN used the geometry of the grid to restrict room connectivity. However, this dungeon also has an interesting diversity of room layouts. It is possible that the selection pressure to develop interesting room connectivity patterns inserted so many diverse activation functions into the the CPPNs that more diverse room patterns emerged as a side-effect. The Direct2GAN dungeon (Fig.\ \ref{fig:dungeonDirectToGAN}) also features diverse room patterns, but they are not arranged in a cohesive fashion, and this approach is also more susceptible to creating isolated, disconnected rooms.

\section{Conclusions and Future Work}
Generative Adversarial Networks (GANs) have shown impressive results as generators for high-quality images. However, combining multiple GAN-generated patterns into a cohesive whole, which is especially relevant for level generation in games, was so far an unexplored area of research. This paper presents the first method that can create large-scale game levels through the combination of a Compositional Pattern Producing Network and a GAN. One of the main insights is that there is a functional relationship between the latent vectors of different game segments, which the CPPN is able to exploit. Additionally, in comparison to a direct representation of multiple latent vectors, the introduced CPPN2GAN approach was able to generate a larger variety of different complete game levels.

However, CPPN2GAN results often contained repeated segments. If this is undesirable for different applications, 
future experiments could use measures for solution variety that also consider local diversity.
Such measures might do a better job of encouraging repetition \emph{with} variation instead of pure repetition.

Beyond our proposed approach, there are other interesting ways to combine the two techniques in this paper. For example, CPPNs could be used to generate global structure only, while modules are evolved independently to fulfill local objectives. For Zelda, this would mean that the global dungeon structure (room presence and their connections) is generated via CPPN, but the separate rooms are determined by evolved real vectors. Such a combination would create interesting global layouts, but allow for arbitrary, chaotic variation when going from room to room.

The two approaches could also be combined sequentially. First, CPPN2GAN would introduce some initial global structure and impose some regular patterns in the level. Afterwards, Direct2GAN could take over to fine-tune the results and/or introduce more local variety. This idea is similar to HybrID \cite{CluneBPO09}. 



However, these extensions lose the ability to scale to arbitrary sizes; a major benefit of CPPNs. In fact, the CPPN2GAN encoding could enable levels in which components are lazily generated as needed. This would be especially useful for exploration games, as new modules (e.g.\ dungeon rooms) can be served by CPPN2GAN whenever new areas of the map are discovered by the player.


Besides different combination approaches, modifications to the training process are also possible. Here, the GAN was pre-trained and only the CPPNs were evolved.
However, it should be possible to make use of a discriminator to also decide whether global patterns (such as Zelda room structure) are similar to the original levels. This would mean adversarial training of the complete CPPN2GAN network against a discriminator. 
Training could be end-to-end or by alternating between the CPPN and the generator. The resulting samples should be able to reproduce both global \emph{and} local patterns in complete game levels rather than just in individual segments. 




\begin{acks}

The authors would like to thank the Schloss Dagstuhl team and
the organisers of Dagstuhl Seminars 17471 and 19511 
for hosting productive seminars.

\end{acks}

\bibliographystyle{ACM-Reference-Format}
\bibliography{CPPNtoGAN} 


\begin{figure*}
\centering
\begin{subfigure}{0.49\textwidth} 
    \includegraphics[width=1.0\textwidth]{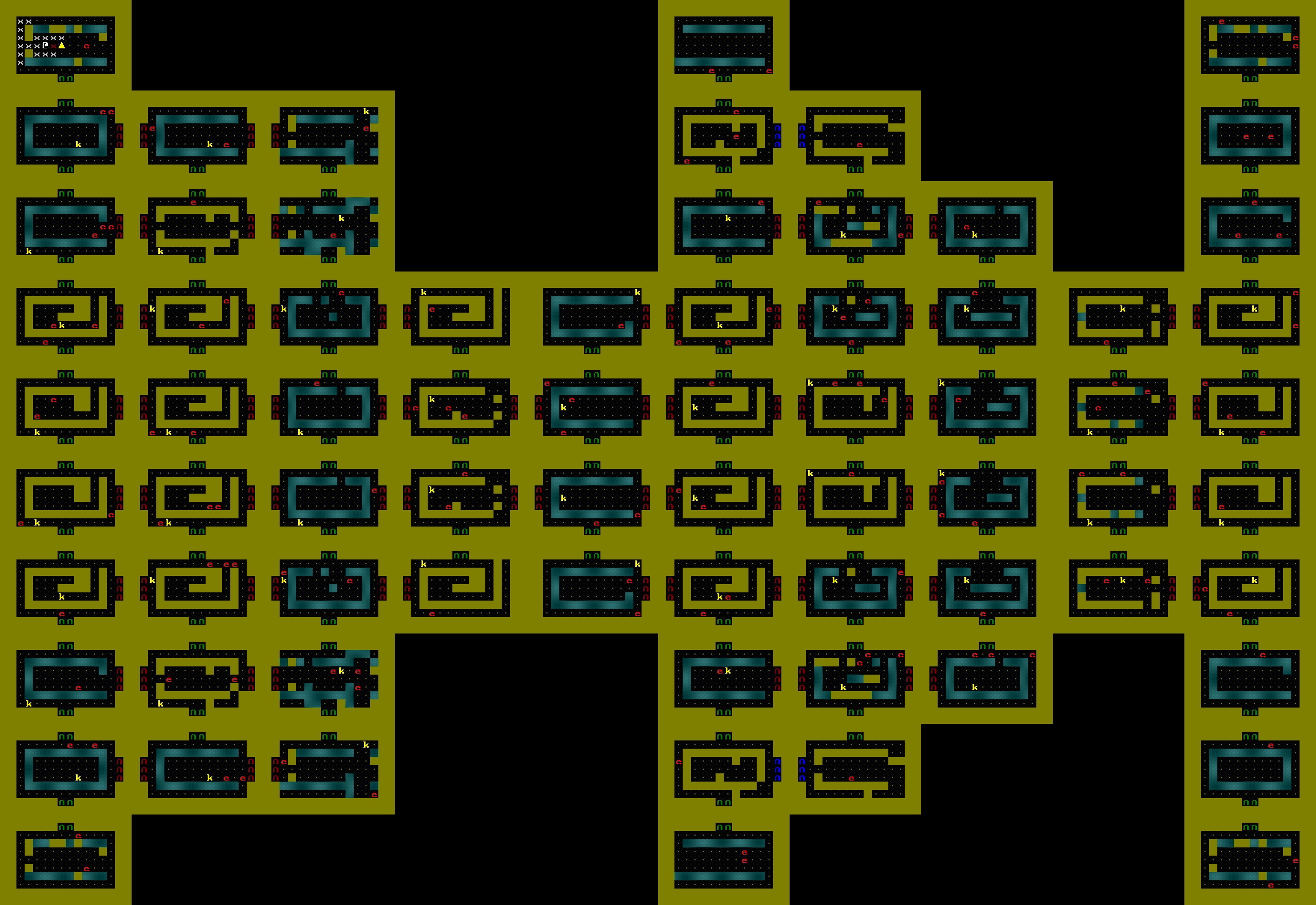}
\end{subfigure}
\begin{subfigure}{0.49\textwidth}
    \includegraphics[width=1.0\textwidth]{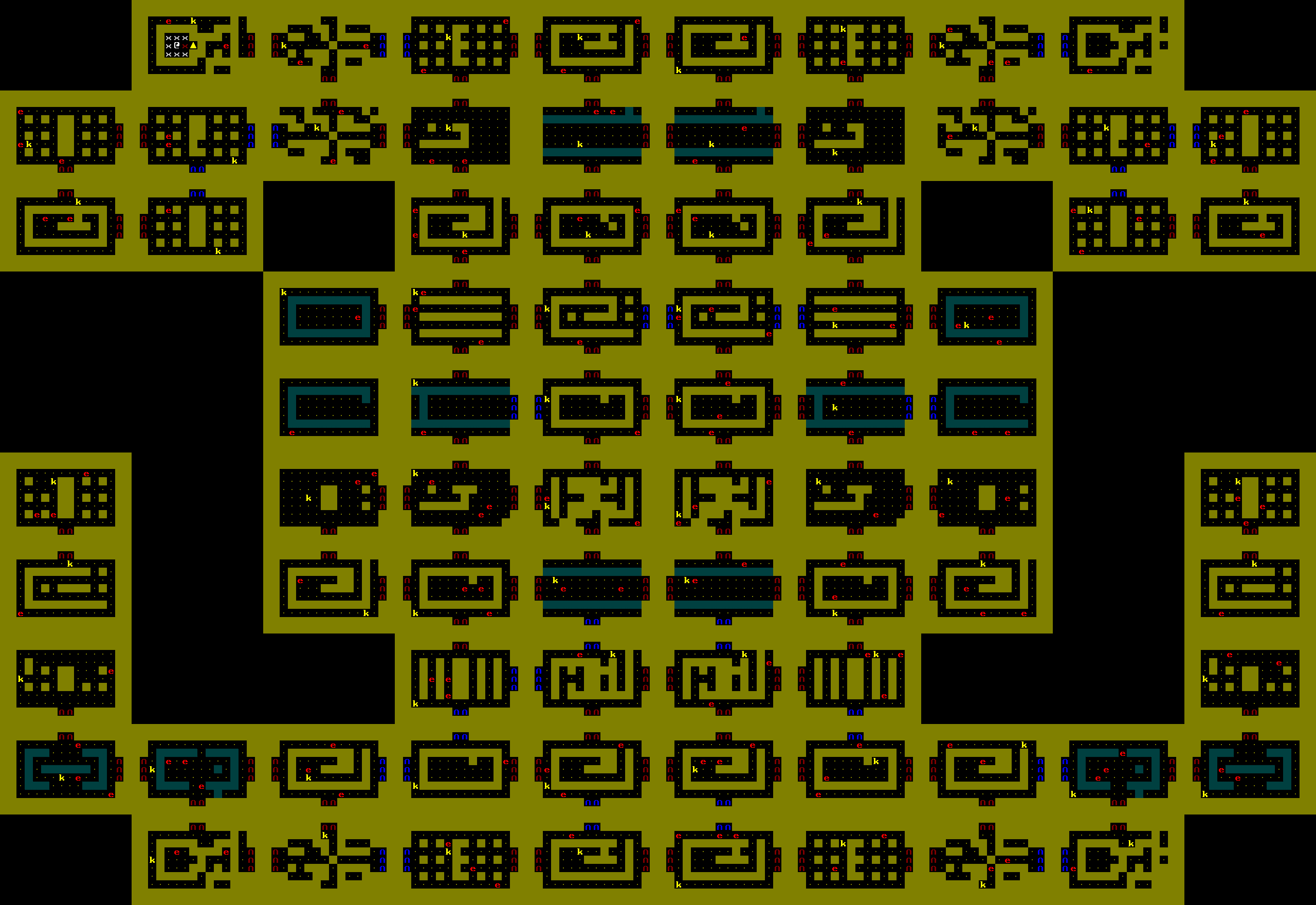}
\end{subfigure}
\begin{subfigure}{0.49\textwidth}
    \includegraphics[width=1.0\textwidth]{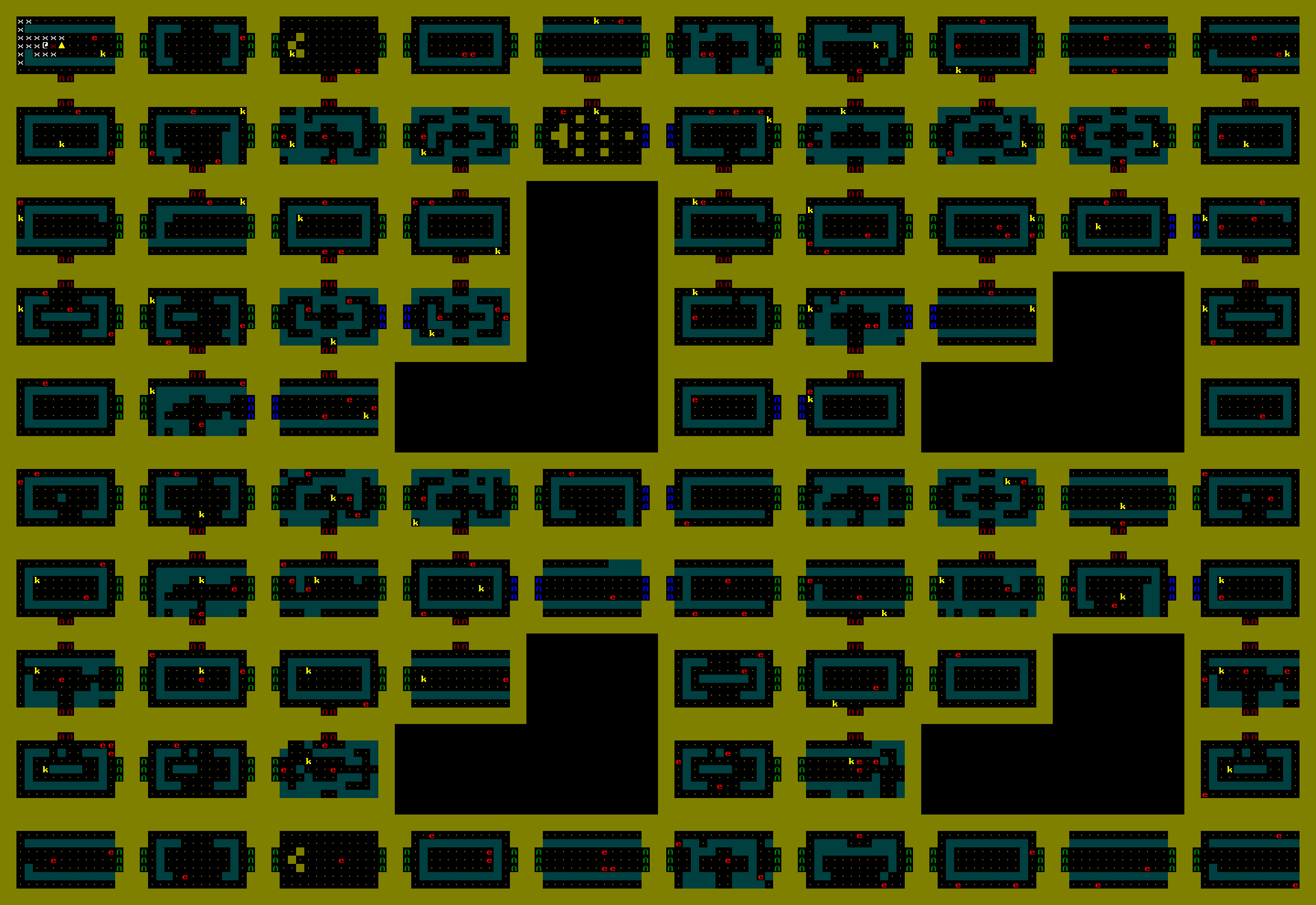}
\end{subfigure}
\begin{subfigure}{0.49\textwidth}
    \includegraphics[width=1.0\textwidth]{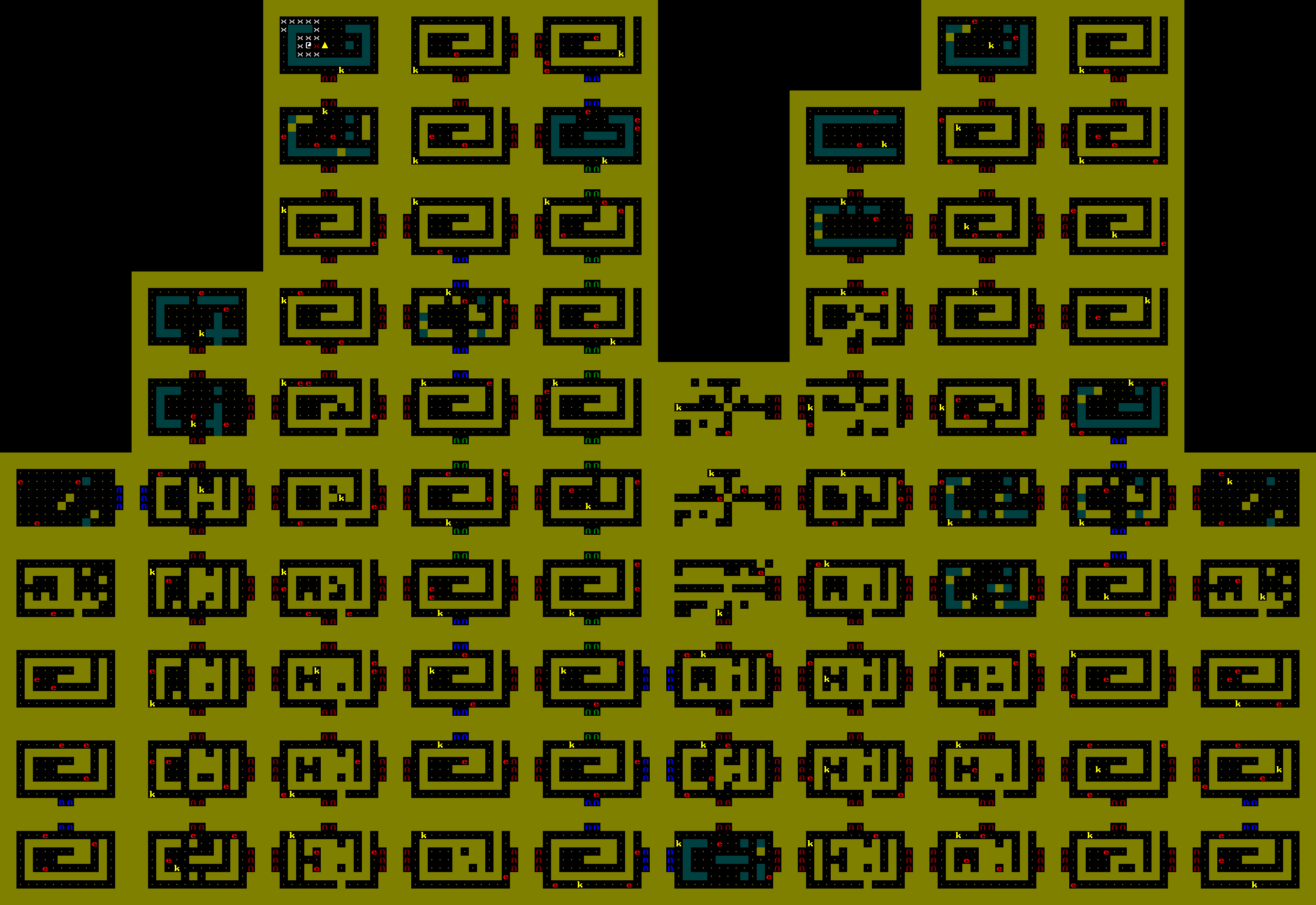}
\end{subfigure}
\begin{subfigure}{0.49\textwidth}
    \includegraphics[width=1.0\textwidth]{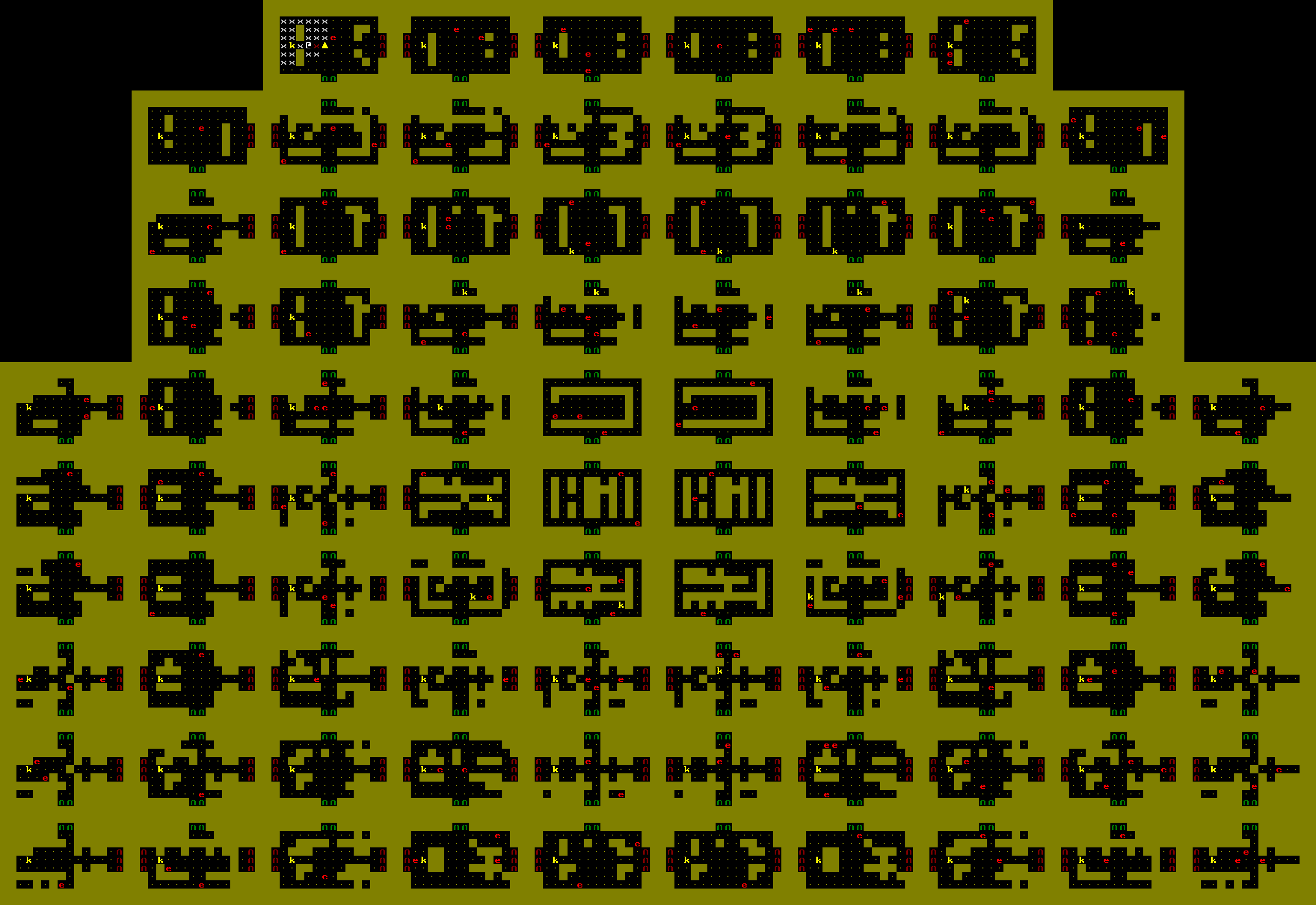}
\end{subfigure}
\begin{subfigure}{0.49\textwidth}
    \includegraphics[width=1.0\textwidth]{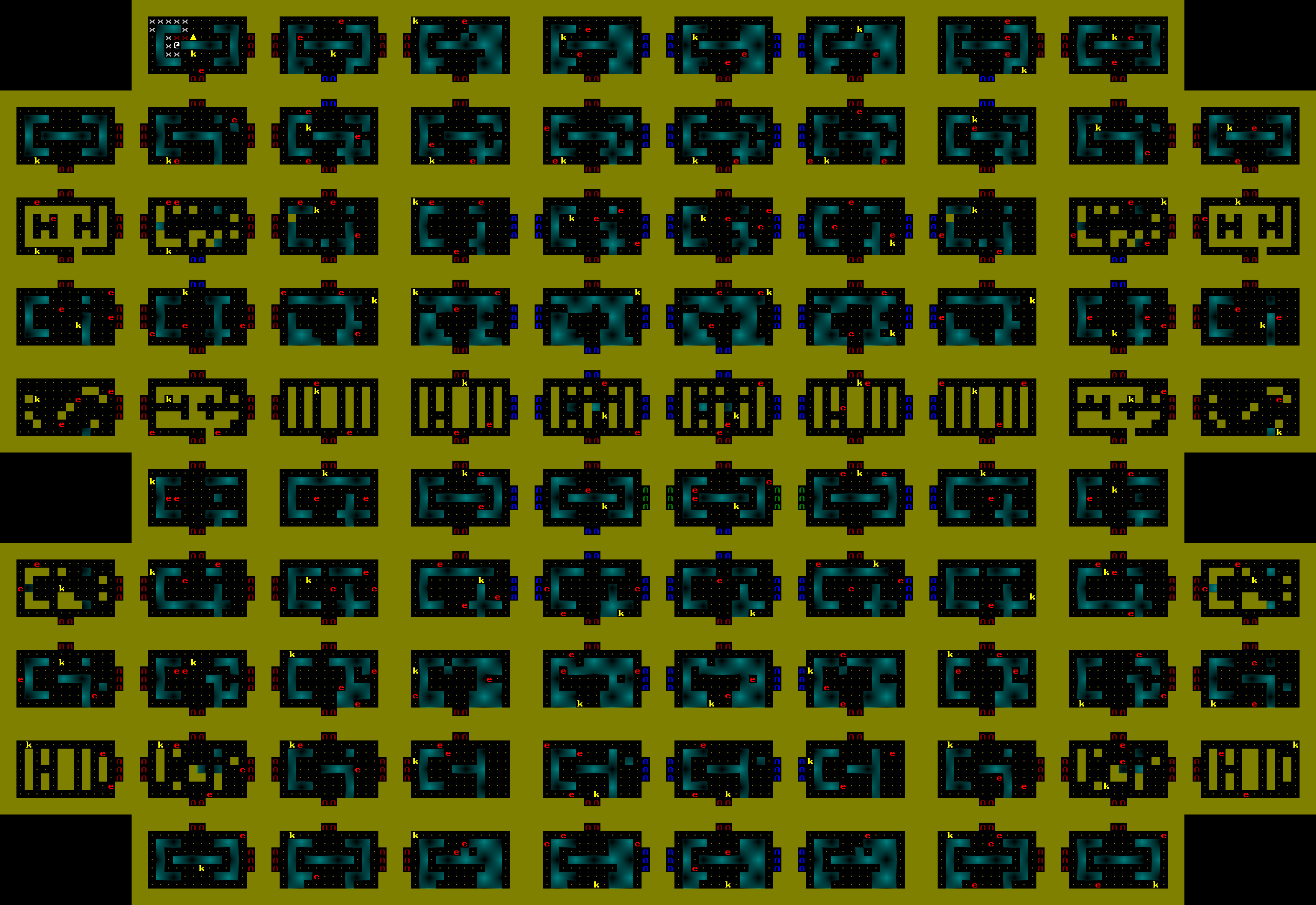}
\end{subfigure}
\caption{A selection of example Zelda levels created by CPPN2GAN. \normalfont The maps show varying degrees of symmetry, repetition, and repetition with variation.}
\label{fig:cppn}
\end{figure*}


\begin{figure*}[t]
\centering
    \includegraphics[width=1.0\textwidth]{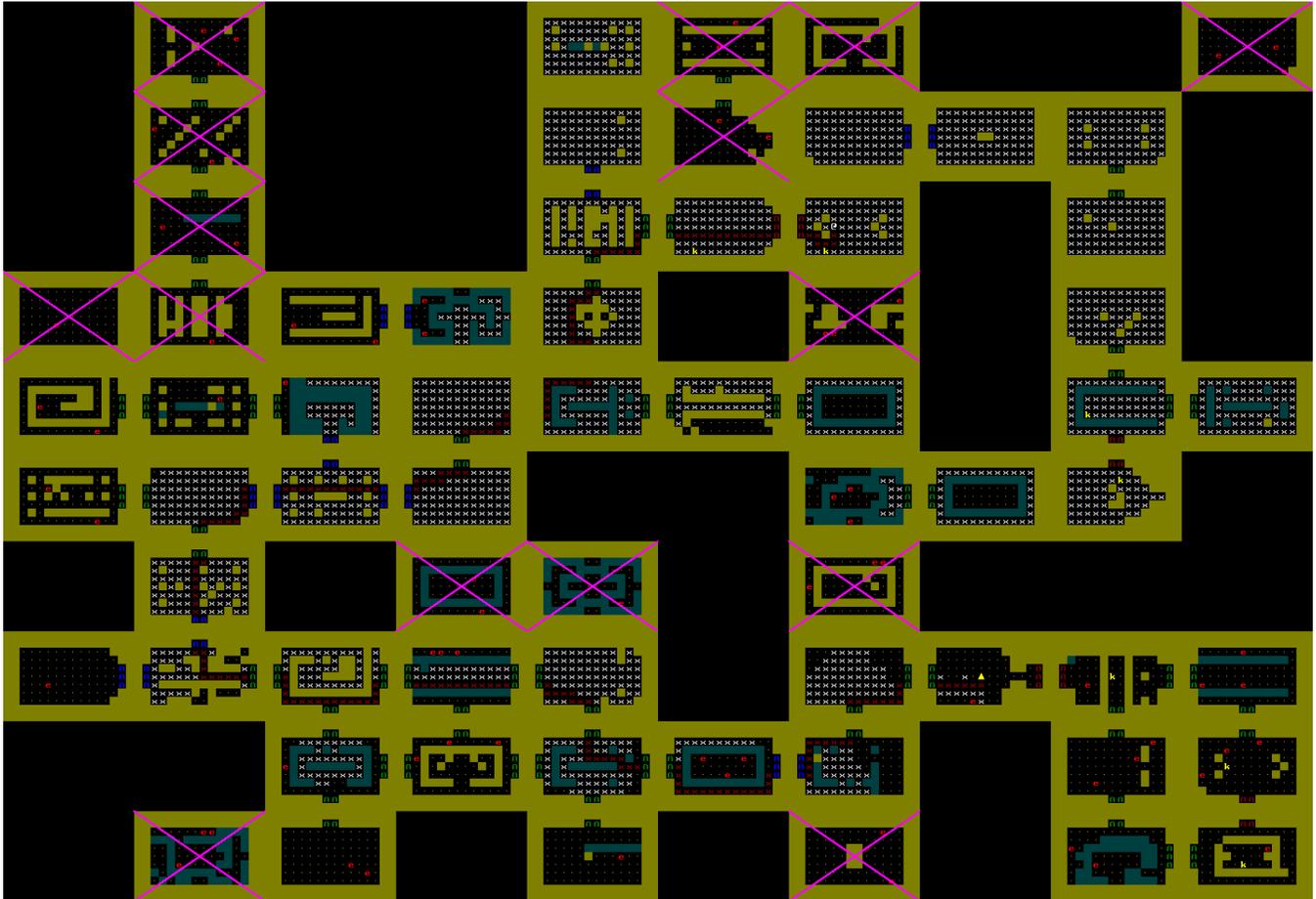}
    \caption{Direct2GAN Dungeon: 50 Reachable Rooms (large version of Fig.~\ref{fig:dungeonDirectToGAN}).
    \normalfont
The @ symbol is the start point, and the triangle is the goal. The red X marks from the start to the goal mark the solution path, and the white X marks correspond to locations checked by A* search in order to find the solution. 
The Direct2GAN dungeon is more sprawling, and has several rooms that are not reachable (\textcolor{magenta}{magenta X}). CPPN2GAN dungeons are more cohesive, and themes can be noticed in different regions of the dungeons.}

    \label{bigfig:dungeonDirectToGAN}
\end{figure*}

\begin{figure*}[t]
    \includegraphics[width=1.0\textwidth]{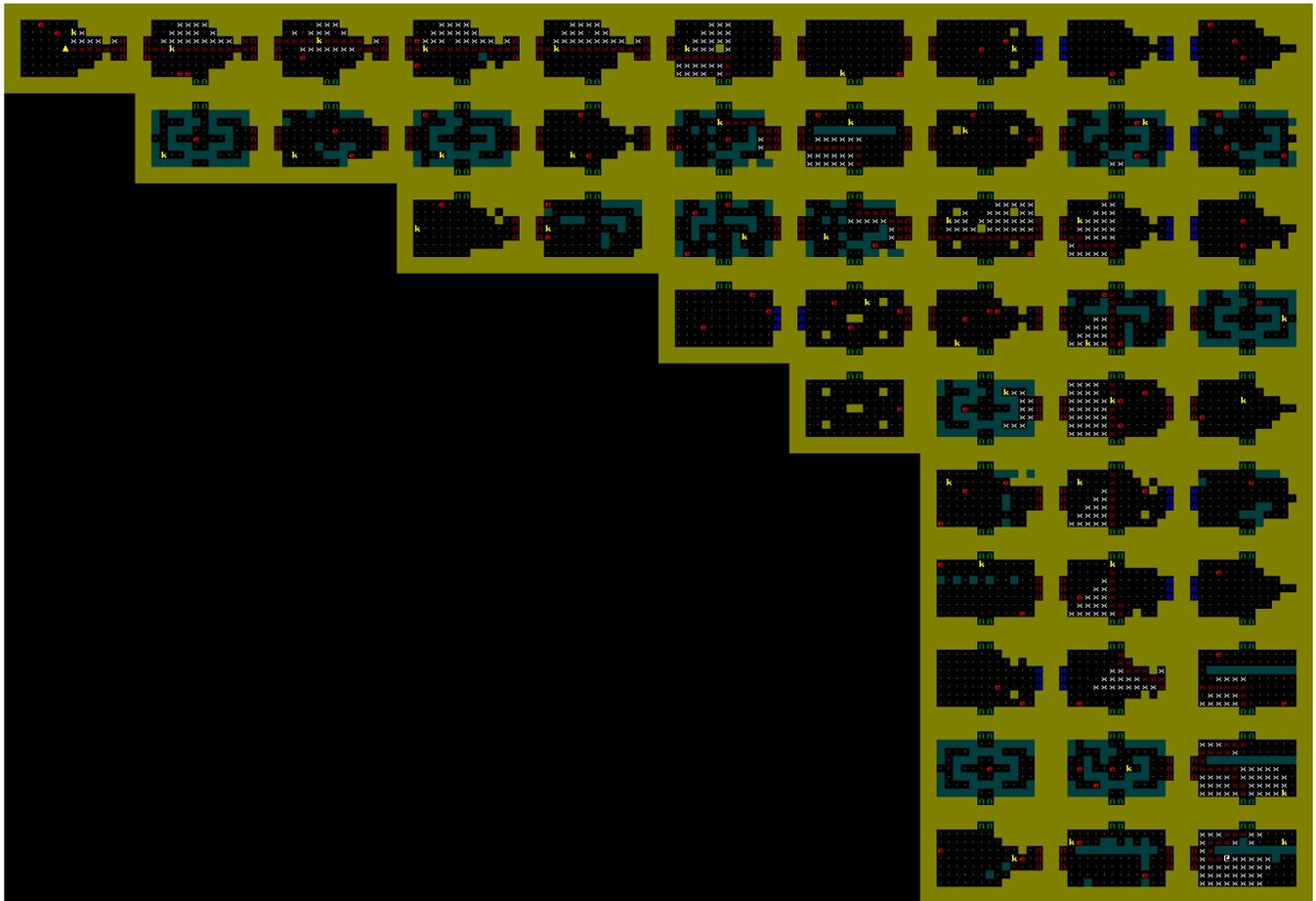}
    \caption{CPPN2GAN: 50 Reachable Rooms (large version of Fig.~\ref{fig:dungeonCPPNtoGAN}).
    \normalfont
The @ symbol is the start point, and the triangle is the goal. The red X marks from the start to the goal mark the solution path, and the white X marks correspond to locations checked by A* search in order to find the solution. 
CPPN2GAN dungeons are more cohesive, and themes can be noticed in different regions of the dungeons.}
    \label{bigfig:dungeonCPPNtoGAN}
\end{figure*}

\begin{figure*}[t]
    \includegraphics[width=1.0\textwidth]{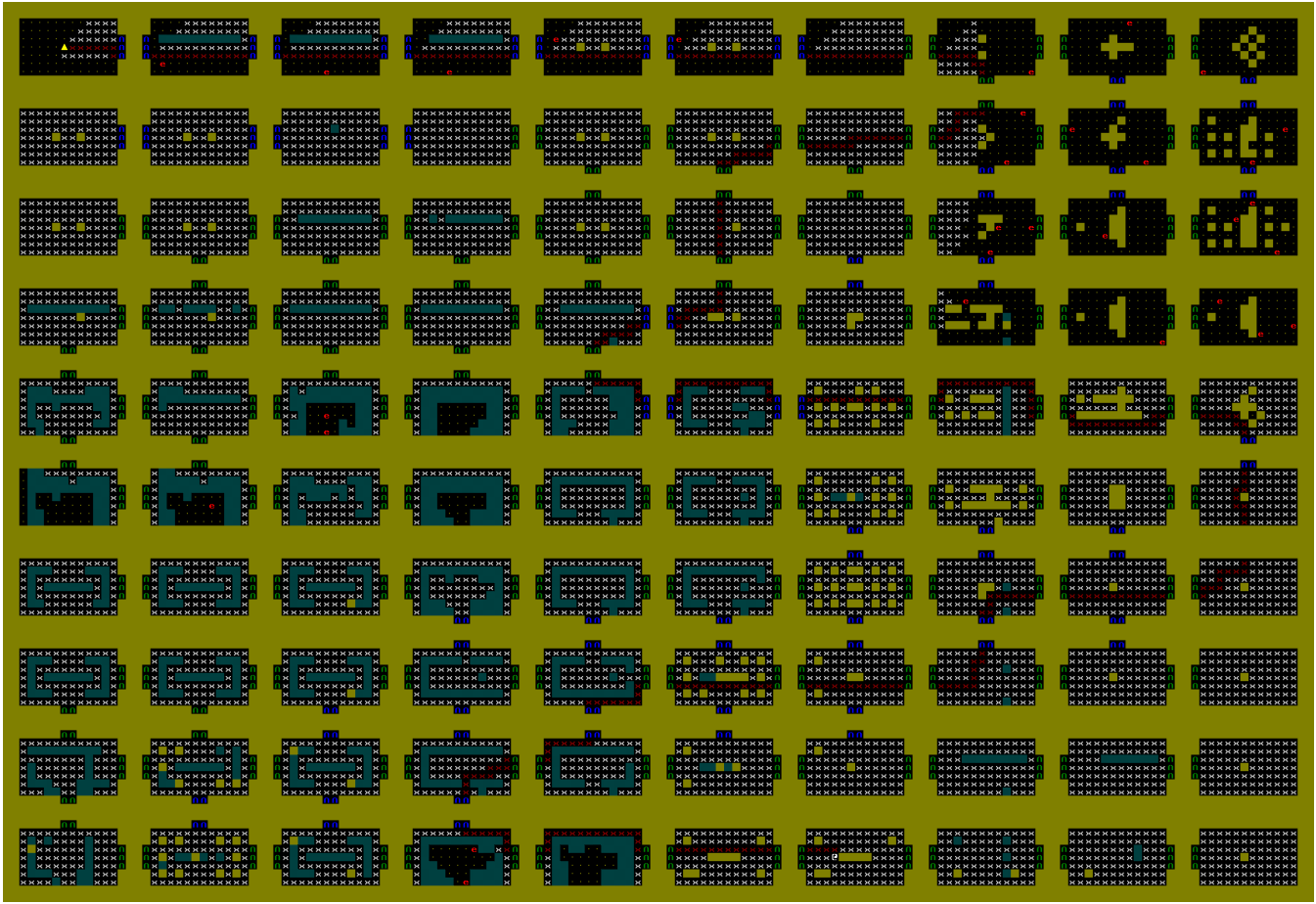}
    \caption{CPPN2GAN: 100 Reachable Rooms (large version of Fig.~\ref{fig:dungeonCPPNtoGAN100}).
    \normalfont
The @ symbol is the start point, and the triangle is the goal. The red X marks from the start to the goal mark the solution path, and the white X marks correspond to locations checked by A* search in order to find the solution. 
CPPN2GAN dungeons are more cohesive, and themes can be noticed in different regions of the dungeons.}
    \label{bigfig:dungeonCPPNtoGAN100}

\end{figure*}

\end{document}